\documentclass[final,5p,times,twocolumn]{elsarticle}
\usepackage[utf8]{inputenc}

\usepackage[linesnumbered]{algorithm2e}
\SetKw{Yield}{yield}
\DontPrintSemicolon
\SetKwFor{ForEach}{foreach}{}{}
\SetKwIF{If}{ElseIf}{Else}{if}{}{else if}{else}{}
\SetKwProg{Fn}{function}{}{} 
\AlgoDontDisplayBlockMarkers

\SetKwFunction{MineType}{MineType}%
\SetKwFunction{MineValue}{MineValue}%
\SetKwFunction{MineSelf}{MineSelf}%
\SetKwFunction{MineEnum}{MineEnum}%
\SetKwFunction{MineDatatype}{MineDatatype}%
\SetKwFunction{MineSome}{MineSome}%
\SetKwFunction{SLDM}{SLDM}%
\SetKwFunction{BuildIndex}{BuildIndex}%
\SetKwFunction{PruneIndex}{PruneIndex}%
\SetKwFunction{MineClosedConjunctions}{MineClosedConjunctions}

\usepackage{amsmath}
\usepackage{amssymb}
\usepackage{alltt}
\usepackage{forest}

\usepackage{float}
\newfloat{listing}{thp}{lop}
\floatname{listing}{Listing}

\usepackage{tikz}
\usetikzlibrary{calc}
\usepackage{pgfplots}
\usetikzlibrary{trees}
\usetikzlibrary{positioning}
\pgfplotsset{every tick label/.append style={font=\small}}

\newcommand{\subjects}[3]{
\begin{scope}[name prefix=#1,node distance=0cm]
\node (0) at #2 {};
\foreach \l [count=\n,remember=\n as \prev (initially 0)] in #3
{
	\node[subject,right=of \prev] (\n)  {\texttt{\l}};
}
\end{scope}
}

\usepackage{url}
\usepackage{alltt}
\usepackage{tabtt}
\usepackage{multirow}
\usepackage{subcaption}  
\usepackage{array}
\usepackage{placeins}

\usepackage[utf8]{inputenc}
\newcommand{\typedliteral}{\textasciicircum\textasciicircum}
\newcommand{\owl}[1]{\ensuremath{\,\textsc{#1}\,\,}}
\newcommand{\owlAnd}{\owl{and}}
\newcommand{\owlSome}{\owl{some}}

\newcommand{\owlValue}{\owl{value}}
\newcommand{\owlSelf}{\owl{Self}}
\newcommand{\owlSubClassOf}{\owl{SubClassOf:}}
\DeclareMathOperator{\support}{\sigma}
\newcommand{\graph}{\mathbb{G}}
\DeclareMathOperator{\match}{\mu}
\newcommand{\type}{\texttt{rdf:type}}
\newcommand{\uris}{\ensuremath\mathcal{U}}
\newcommand{\literals}{\ensuremath\mathcal{L}}
\newcommand{\weight}{w}
\newcommand{\len}{\ensuremath\mathfrak{d}}
\newcommand{\Patterns}{\ensuremath\mathcal{P}}
\newcommand{\triples}{\ensuremath\mathcal{T}}
\newcommand{\Index}{\ensuremath\mathcal{I}}

\newcommand{\Hobbit}{\texttt{:The\_Hobbit}}
\newcommand{\Silmarillion}{\texttt{:The\_Silmarillion}}
\newcommand{\FotR}{\texttt{:The\_Fellowship\_of\_the\_Ring}}
\newcommand{\TT}{\texttt{:The\_Two\_Towers}}
\newcommand{\RotK}{\texttt{:The\_Return\_of\_the\_King}}

\newcommand{\sparql}{SPARQL}
\newcommand{\shacl}{SHACL}
\newcommand{\name}[1]{\emph{#1}}
\newcommand{\protege}{\name{Prot\'eg\'e}}

\newcommand{\transform}{\ensuremath\mathbf{\tau}}

\RequirePackage{pifont}
\newcommand{\cmark}{\text{\ding{51}}}

\journal{Journal of Web Semantics}

\begin{document}

\begin{frontmatter}



\title{Swift Linked Data Miner: Mining OWL 2 EL Class Expressions\\Directly from Online RDF Datasets}
\author[put]{J.~Potoniec\corref{cor}}
\ead{jpotoniec@cs.put.poznan.pl}
\cortext[cor]{Corresponding author}
\author[put]{P.~Jakubowski}
\ead{pjakubowski@cs.put.poznan.pl}
\author[put]{A.~Ławrynowicz}
\ead{alawrynowicz@cs.put.poznan.pl}

\address[put]{Faculty of Computing, Poznan University of Technology, ul.~Piotrowo~3, 60-965~Poznań, Poland}

\newdefinition{definition}{Definition}


\begin{abstract}
In this study, we present Swift Linked Data Miner, an interruptible algorithm that can directly mine an online Linked Data source (e.g., a SPARQL endpoint) for OWL 2 EL class expressions to extend an ontology with new \owlSubClassOf axioms.
The algorithm works by downloading only a small part of the Linked Data source at a time, building a smart index in the memory and swiftly iterating over the index to mine axioms.
We propose a transformation function from mined axioms to RDF Data Shapes.
We show, by means of a crowdsourcing experiment, that most of the axioms mined by Swift Linked Data Miner are correct and can be added to an ontology.
We provide a ready to use \protege{} plugin implementing the algorithm, to support ontology engineers in their daily modeling work.
\end{abstract}


\end{frontmatter}


\section{Introduction}
\label{}

It is reasonable to assume that a Resource Description Framework (RDF \cite{Wood:14:RCA}) graph is not a random set of triples.
There is some reason for the graph to have this specific shape: some underlying data model or a process generating the graph.
An element of the model is reflected as a pattern in the graph.
The idea of Swift Linked Data Miner (SLDM) is to make use of this observation backward: if there is a pattern in the graph, then there is a good amount of chance that there should be a corresponding element in the model.

One of possible ways to express such a model is to generate an ontology.
To express an ontology for an RDF graph, one can employ RDF Schema \cite{rdfs} or aim for some variant of OWL 2 (Web Ontology Language \cite{owl2primer}).
RDF Schema offers just a tiny bit of expressive power, so we decided to choose OWL 2 instead.
One may argue that OWL 2 has exactly the opposite problem, as it is a very expressive language and thus reasoning complexity is unbearable.
Fortunately, there are three OWL 2 profiles: OWL 2 QL, OWL 2 RL, and OWL 2 EL, all of these provide polynomial-time reasoning algorithms.
According to the OWL 2 specification \cite{owl2_overview}, the first two are suitable for lightweight ontologies.
However, OWL 2 EL is a profile tailored specifically to deal with very large ontologies.
It is our intuition that one can easily get a large ontology if they do not have to develop it by hand, but has it data mined instead.
It must be noted here that the OWL Lite profile, defined for OWL \cite{owl1} and inherited by OWL 2 has exponential reasoning complexity \cite{owllite_complexity}.
Taking all these into account, we decided that our primary way to express the model is an OWL 2 EL ontology, which we introduce in Section \ref{sec:preliminaries}.

In the age of Big Data, we cannot assume anymore that we can have a direct access to a graph all the time and fit the whole graph to the RAM.
We must be able to process the graph chunk at a time,  sample it, and  retrieve the chunks from a remote location.
Swift Linked Data Miner was developed with all this in mind.
In Section \ref{sec:index}, we show how we retrieve only a part of an RDF graph at a time and organize the retrieved part into a smart structure to facilitate pattern mining.

In Section \ref{sec:sldm}, we describe SLDM, which can mine an RDF graph to extend an OWL 2 EL ontology.
SLDM is composed of many small algorithms, each fitted to mine particular type of patterns.
Section \ref{sec:complexity} provides a theoretical analysis of the algorithm, including both analyses of memory complexity and worst-case computational complexity.

Some users may not be interested in reasoning.
Maybe the logical inference in their application is unnecessary and instead they would like to understand  their graph better or validate if a new graph they just received is compatible with the model.
Especially to address these issues there is an ongoing work on RDF Data Shapes\footnote{\url{https://www.w3.org/2014/data-shapes/charter}}, and we provide an alternative way to express patterns mined with SLDM.
In Section \ref{sec:shapes}, we show how the patterns can be transformed to RDF Data Shapes expressed in Shapes Constraint Language (\shacl{}) \cite{shacl}.

To facilitate incremental research and enable early adopters to use our ideas, we provide an implementation of SLDM.
In Section \ref{sec:protege}, we present a plugin to \protege{}, which enables ontology engineer to use SLDM in just few clicks.

To validate SLDM, we planned and conducted a crowdsourcing experiment using arguably the most popular Linked Data resource: \name{DBpedia} \cite{dbpedia}.
We used SLDM to mine patterns for multiple classes used in \name{DBpedia}, translated the patterns to English and asked the contributors of a crowdsourcing platform if these sentences correctly describe the classes.
They decided that most of them are indeed correct.
Details of the experiment are described in Section \ref{sec:dbpedia}.

In Section \ref{sec:myexperiment}, we show another use-case, based on myExperiment RDF dataset and the associated ontology \cite{myexperiment}.
We compare the mined patterns with documentation and pragmatics of the dataset.
Finally, in Section \ref{sec:further_exp}, we discuss run-time properties of the algorithm, such as CPU time in function of various parameters of the algorithm.

\section{Related work}

The idea of automated and semi-automated creating and extending ontologies has been studied by many researchers.
These studies can be roughly divided into five areas: (i) ontology learning from text, (ii) interactive ontology learning, (iii) concept learning, (iv) learning from local data, (v) learning from online data.
The first area is also the farthest from our work, and thus we will not discuss it.
The interested reader is refereed to a comprehensive compendium on ontology learning \cite{lehmann2014perspectives}.

One of the prominent ways of interactive ontology learning is an application of formal concept analysis to the Description Logics \cite{DBLP:conf/ijcai/BaaderGSS07,DBLP:conf/icfca/Rudolph08,DBLP:conf/ekaw/Potoniec14,DBLP:conf/semweb/PotoniecRL14}.
The aim of these algorithms is to complete an ontology with respect to  all subsumptions of a given type, for example, between named classes \cite{DBLP:conf/ijcai/BaaderGSS07} or between named classes and domain and range restrictions \cite{DBLP:conf/icfca/Rudolph08}.
The algorithms generate all possible subsumption axioms that are consistent with the ontology but do not follow from it, and for each axiom, the user is asked to either add the axiom to the ontology or to provide a counterexample.

In the same area fits the idea of games with a purpose, where a player receives rewards for completing tasks like creating new entities in an ontology, adding types or aligning an ontology with another one.
The consensus is obtained by posing the same task to multiple players.
A classical example of such a system are \emph{OntoGames} \cite{DBLP:journals/expert/SiorpaesH08}.
Similar to the games with a purpose is the idea of using crowdsourcing services like \emph{CrowdFlower}\footnote{\url{http://www.crowdflower.com}} or \emph{Amazon Mechanical Turk}\footnote{\url{http://www.mturk.com}}.
For example, Hanika et al. present a \protege{} plugin that enables a user to post microtasks related to her ontology development process directly to a crowdsourcing website \cite{DBLP:conf/ekaw/HanikaWS14}.

Concept learning by itself is not necessarily an ontology learning approach, as it is concerned with learning class expressions given a set of positive and negative examples.
Nevertheless, such an approach can be used to extend an existing ontology with missing definitions.
Fanizzi et al.  propose DL-FOIL algorithm, which allows for learning class expressions in Description Logics underlying OWL-DL \cite{DBLP:conf/ilp/FanizzidE08,Harmelen:04:OWO}.
The learning algorithm is based on sequential covering and employs two refinement operators, one for specialization and the other  for generalization.
Lehmann  describes a concept learning software \emph{DL-Learner}\footnote{\url{http://dl-learner.org}}, which also employs refinement operators, but with different search strategy \cite{jl_2009/dllearner_jmlr}.
He also describes the  application of \emph{DL-Learner} specifically to learning ontologies, along with a \protege{} plugin implementing the idea \cite{DBLP:journals/ws/LehmannABT11}.
An approach using a refinement operator with background knowledge to refine SPARQL queries was proposed by Ławrynowicz and Potoniec  \cite{DBLP:journals/ijswis/LawrynowiczP14}.
These queries are further used as binary features for a classical machine-learning classification algorithm.
Due to certain properties of these SPARQL queries, they can be immediately transformed to OWL class expressions, and thus also used in ontology learning, as discussed in  a previous study \cite{DBLP:conf/aaai/PotoniecL15}.
Gal{\'{a}}rraga et al.  employ techniques known from Inductive Logic Programming on top of their in-memory RDF store  to mine association rules with variables \cite{DBLP:conf/www/GalarragaTHS13}.
Such rules could be then transformed into an ontology.

The idea of learning ontological axioms from a static, fully-available dataset is  related to various forms of data mining in relational databases.
For example, Fu and Han  propose a framework for mining association rules in relational databases guided by constrains on a shape of mined rules \cite{DBLP:conf/dood/FuH95}.
By setting these constrains appropriately, one could obtain an ontology.
V{\"{o}}lker et al. propose an algorithm for mining an OWL 2 EL ontology from scratch \cite{DBLP:conf/otm/FleischhackerV11, DBLP:conf/esws/VolkerN11}.
The input to the algorithm is an RDF graph, which is first transformed into a set of database tables, and then association rule mining is employed to discover the ontological axioms.
Instead of considering an RDF graph and an ontology describing it, one could tackle the problem of extending an ontology using individuals contained in it. 
Another study provides a method for learning general class inclusions (GCIs) that takes into account results of reasoning with the ontology \cite{DBLP:conf/semweb/SazonauSB15}.

Finally, one may want to extend an ontology using data that are not entirely available at hand but   are available only in for example, remote SPARQL endpoints \cite{sparql}.
To the best of our knowledge, there is  the only one work that has addressed this problem so far \cite{DBLP:conf/semweb/BuhmannL13}.
It is a top-down method, which first performs data mining on a repository of ontologies to build a library of patterns and then the patterns are used to form SPARQL queries, which are posed to  a SPARQL endpoint to discover axiom candidates.
Unfortunately, the queries are computationally expensive for the endpoint, due to the heavy usage of \texttt{\textsc{group by}} and \texttt{\textsc{count distinct}} clauses.

\section{Preliminaries\label{sec:preliminaries}}
\subsection{OWL 2 EL\label{sec:owl2el}}

OWL 2 is a language designed to describe information about entities and relations between them.
The language provides formally defined semantics and decidable reasoning procedures \cite{owl2primer}.
OWL 2 EL is a subset of OWL 2, tailored to support applications employing very large ontologies while having typical reasoning tasks tractable \cite{owl2profiles}.
For example, this subset is used by a large clinical health ontology SNOMED CT \cite{snomedct}.

Throughout this work we write OWL expressions in Manchester Syntax \cite{manchester}.
We also use a set of well-known prefixes: \texttt{rdf:} for the namespace \url{http://www.w3.org/1999/02/22-rdf-syntax-ns#}, \texttt{rdfs:} for \url{http://www.w3.org/2000/01/rdf-schema#}, \texttt{owl:} for \url{http://www.w3.org/2002/07/owl#} and \texttt{xsd:} for \url{http://www.w3.org/2001/XMLSchema#}.

We start by defining an \emph{OWL 2 EL class expression}.
Let $A$ be a named class, $DT$ a datatype, $p$ a named object property, and $r$ a named data property.
Moreover, $a$ denotes an individual and $l$ a literal.
Following the study of Motik et al.  \cite{owl2profiles}, the OWL 2 EL class expression $C^{EL}$ is defined as follows:

\begin{align*}
	C^{EL} := & A | C^{EL} \owlAnd C^{EL} | \{a\} | p \owlSome C^{EL} | p \owlValue a | 
	p \owlSelf | \\&  r \owlSome R^{EL} | r \owlValue l \\
	R^{EL} := & DT | R^{EL} \owlAnd R^{EL} | \{l\} \\
\end{align*}

The datatypes in OWL 2 EL are severely limited compared to the full OWL 2.
There are only 19 datatypes that were chosen such that their intersections are either infinite or empty \cite{owl2profiles}.
Magka et al. proved that OWL 2 EL class expressions can be extended with inequalities over numeric domains \cite{Magka2011}.
For a \owlSubClassOf axioms with a subclass expression in OWL 2 EL, a superclass expression can be extended with $\geq$ and $\leq$ relations over real, rational, and integer numbers, and with $\geq$ relation over natural numbers.
Following the idea and using the same symbols as before, we can define the \emph{OWL 2 EL superclass expression} $C$ as follows:
\begin{align*}
	C := & A | C \owlAnd C | \{a\} | p \owlSome C | p \owlValue a | p \owlSelf | \\ 
	& r \owlSome R | r \owlValue l \\
	R := & DT | R \owlAnd R | \{l\} | LT[<= l] | GT[>= l] \\
	LT := & \texttt{owl:real} | \texttt{owl:rational} | \texttt{xsd:decimal} | \texttt{xsd:integer} | \\
	     & \texttt{xsd:dateTime} | \texttt{xsd:dateTimeStamp} \\
	GT := & LT | \texttt{xsd:nonNegativeInteger}
\end{align*}
According to the specification \cite{xml_datatypes}, types \texttt{xsd:dateTime} and \texttt{xsd:dateTimeStamp} closely correspond to decimal numbers, so we can safely incorporate them in the definition.

\subsection{Datatypes in OWL\label{sec:datatypes}}

\begin{figure}
\resizebox{\columnwidth}{!}
{
\begin{forest}
for tree={grow=south
,edge path={%
            \noexpand\path [\forestoption{edge}] (!u.parent anchor) -- +(0,-15pt) -| (.child anchor)\forestoption{edge label};
        },},
[\texttt{rdfs:Literal}
[\texttt{owl:real} [\texttt{owl:rational} [\texttt{xsd:decimal} [\texttt{xsd:integer} [\texttt{xsd:nonNegativeInteger}]]]]]
[\texttt{xsd:string} [\texttt{xsd:normalizedString} [\texttt{xsd:token},tier=l3 [\texttt{xsd:NMTOKEN},tier=l4] [\texttt{xsd:Name} [\texttt{xsd:NCName},tier=l5]]]]]
[\texttt{xsd:dateTime} [\texttt{xsd:dateTimeStamp},tier=l2]]
]
\end{forest}
}
\caption{The hierarchy of datatypes in OWL 2 EL.
For clarity, omitted are direct children of \texttt{rdfs:Literal} with no children themselves, that is, \texttt{rdf:PlainLiteral, rdf:XMLLiteral, xsd:hexBinary, xsd:base64Binary, xsd:anyURI}.
\label{fig:datatypes-hierarchy}}

\end{figure}
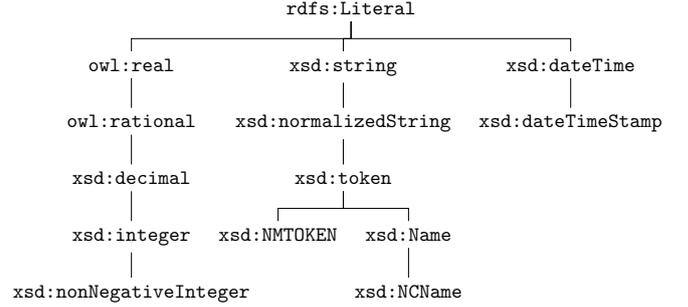

OWL heavily relies on the datatypes defined for XML Schema Definition Language \cite{xml_datatypes}.
These datatypes are organized into such a hierarchy, that a valid value for a subtype is also a valid value for a supertype, for example,  $3$ is a valid value for  \texttt{xsd:nonNegativeInteger}, but also for \texttt{xsd:integer} and all its supertypes, including \texttt{rdfs:Literal}.
In Figure \ref{fig:datatypes-hierarchy}, we present the datatype hierarchy for OWL 2 EL, based on the definitions of the value spaces for the datatypes  \cite{xml_datatypes, rdfs, owl2}.

\section{Three level index for an efficient access to an RDF graph\label{sec:index}}

We aim at data mining directly on Linked Data, that is, we want to be able to operate in a setup with only parts of an RDF graph available at a time.
We consider a \sparql{} endpoint with a very simple \sparql{} queries posed to it, but our approach could be easily adapted to other, less restrictive (in terms of server workload) frameworks available in the Web such as Linked Data Fragments, subject pages or even data dumps  \cite{ldf}.

Precisely, for a given set of Uniform Resource Identifiers (URIs) $\uris$ we build a set of \sparql{} queries based on the following template:
\begin{alltt}
	select ?s ?p ?o
	where
	\{
	    ?s ?p ?o.
	    values ?s \{\ldots\}
	\}
\end{alltt}
We replace \texttt{\ldots} with small disjoint subsets of $\uris$.
In the implementation described in Section \ref{sec:protege}, the number of items in a subset is a configurable parameter with a default value of 100 URIs.
Such a query extracts all triples \texttt{(?s, ?p, ?o)} from the endpoint, but with a restriction that all the bindings for the variable \texttt{?s} must be from the set of URIs listed in the curly braces after the \texttt{values} keyword, that is, from the set that replaced \texttt{\ldots} in the template.
The idea is to avoid posing $\left|\uris\right|$ queries to the endpoint, thus lessening network and CPU load, but keep them simple enough to avoid issues with queries being executed too long.

If the set of URIs $\uris$ is large, the number of queries posed by SLDM may be too large for the used \sparql{} endpoint.
To further decrease its load, it is possible to use sampling.
We propose three different strategies for sampling:
\begin{description}
\item[uniform] The set $\uris$ is shuffled and a fixed number of items (e.g., 1000) is kept, the rest discarded.
In other words, we perform simple random sampling without replacement.
This is the baseline strategy, that does not use any additional knowledge during the sampling and does not pose any SPARQL query to the endpoint.
\item[predicates counting]
As SLDM constructs patterns by detecting features shared by a substantial number of URIs, the richer the description of an URI in the RDF graph, the more patterns it can support.
One of possible measures of description richness is the number of distinct predicates used to describe a particular URI.
First, for every URI the number of distinct predicates occurring in triples with the URI as the subject is counted.
It is relatively easy to obtain these numbers using SPARQL 1.1 queries constructed from the following template:
\begin{alltt}
select ?s (count(distinct ?p) as ?c)
where 
\{ 
    ?s ?p [] . 
    values ?s \{\ldots\}
\}
group by ?s
\end{alltt}
The ellipsis in the \texttt{values} clause is to be filled by splitting the set $\uris$, as described earlier.
The obtained numbers are then normalized by their sum (i.e., now the sum is equal to 1) and the normalized numbers are used as a probability of a particular URI being chosen.
\item[triples counting] This strategy employs a very similar idea, but it uses the number of triples instead of the number of distinct predicates as the measure of description richness.
The following query template is used:
\begin{alltt}
select ?s (count(?o) as ?c)
where
\{
    ?s ?p ?o . 
    values ?s \{\ldots\}
\}
group by ?s
\end{alltt}
Notice that the keyword \texttt{distinct} was removed from the query compared to the previous one.
The obtained numbers are again normalized and used as the probability.
\end{description}
The uniform strategy guarantees reduction of the workload, while the other two strategies do so only if the endpoint is capable of efficient answering of group by queries, which may vary depending on the software running the endpoint.
Using sampling we can significantly reduce the workload, but as with every sampling there is a price: it is possible to introduce mistakes to the mined patterns.
As the counting strategies favor the URIs that are more richly described, we think that using them should mitigate some of the sampling issues in case of an RDF dataset of uneven quality.
We provide experimental comparison of the strategies in Section \ref{sec:strategies_comparison}.

\begin{figure}
\centering
\resizebox{\columnwidth}{!}
{
\begin{tikzpicture}[yscale=.6]
\foreach \x in {0,1,...,2}
{
	\draw (0,-\x-2) rectangle ++(1,1) rectangle ++(.5,-1);	
	\node at ($(0,-\x-2)+(.5,.5)$) {$p_\x$};
}
\foreach \x in {0,1,...,2}
{
	\draw (3,-\x) rectangle ++(1,1) rectangle ++(.5,-1);	
	\node at ($(3,-\x)+(.5,.5)$) {$o_\x$};
}
\foreach \x in {3,4}
{
	\draw (3,-\x-.5) rectangle ++(1,1) rectangle ++(.5,-1);	
	\node at ($(3,-\x-.5)+(.5,.5)$) {$o_\x$};
}
\foreach \x in {5}
{
	\draw (3,-\x-1) rectangle ++(1,1) rectangle ++(.5,-1);	
	\node at ($(3,-\x-1)+(.5,.5)$) {$o_\x$};
}
\foreach \yb/\ye in {0/0,1/3.5,2/6}
{
	\draw[->,thick] ($(1.25,-1*\yb+.5-2)$) .. controls ++(1.5,0) and ++(-1.5,0) .. ($(3,-1*\ye+.5)$);
}
\foreach \y/\start/\stop in {0/0/2, 1/3/3, 2/4/5, 3.5/6/8, 4.5/9/9, 6/10/10}
{
\foreach \x in {\start,...,\stop}
{
	\draw ($(6+\x-\start,-1*\y)$) rectangle ++(1,.75);
	\node at ($(6+\x-\start,-1*\y)+(.5,.5*.75)$) {$s_{\x}$};
}
	\draw[->,thick] ($(4.25,-1*\y+.5)$) .. controls ++(1.5,0) and ++(-1.5,0) .. ($(6,-1*\y+.5)$);
}
\node at (.75, -7) {Level 1};
\node at (3.75, -7) {Level 2};
\node at (6.75, -7) {Level 3};
\draw[dotted] (2, 1.5) -- (2, -7.5);
\draw[dotted] (5.25, 1.5) -- (5.25, -7.5);
\end{tikzpicture}
}
\caption{A sample three level index used in SLDM.
The first level consists of predicates, the second level of objects and the third of subjects.
	Empty rectangles denote pointers to the next level.
There are 11 triples in this index: 6 with predicate $p_0$, 4 with predicate $p_1$ and 1 with predicate $p_2$, namely $(s_{10}, p_2, o_5)$.
\label{fig:index}}
\end{figure}
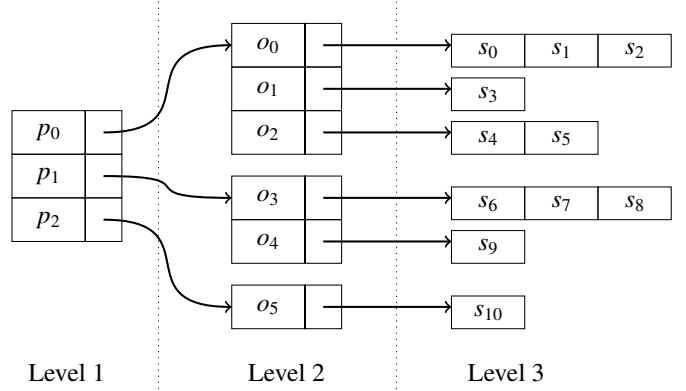

The retrieved triples are stored in a three-level hash-based index structure with predicates in the first level, objects in the second level, and subjects in the third level.
An example of such an index is presented in Figure \ref{fig:index}.
Building the index using hashing is a relatively cheap task: three hash computations for every triple plus amortized constant time insertions.
This particular organization of the index, that is, first predicates, then objects, then subjects, is caused directly by the way our algorithm operates.
Using a different order (e.g., predicates -- subjects -- objects) would increase the computational complexity of the algorithm.
This is because the algorithm operates by grouping together different subjects from triples having the same predicate and object.
The full algorithm for building an index is presented in Algorithm \ref{alg:build_index}.
Its input is a set of triples $\triples$ and it outputs the index $\Index$.

\begin{algorithm}
\Fn{\BuildIndex{$\triples$}}
{
$\Index\leftarrow \emptyset$ \tcp*{An empty three-level index}
\ForEach{$s,p,o\in \triples$}
{
\If{$p\not\in \Index$}
{
\tcp{We encountered a new predicate, initialize data structures for it}
$\Index[p]\leftarrow\emptyset$ \;
}
\If{$o\not\in \Index[p]$}
{
\tcp{We encountered a new object for a predicate $p$}
$\Index[p][o]\leftarrow \emptyset$ \;
}
$\Index[p][o]\leftarrow \Index[p][o]\cup\{s\}$ \;
}
\Return $\Index$
}
\caption{The algorithm building a three-level index for a given set of triples $\triples$.
Square brackets are used as a notation to access the index, for example, $\Index[p]$ refers to a part of the second level of the index, which is pointed by a pointer attached to $p$ in the first level.
\label{alg:build_index}
}
\end{algorithm}

\section{SLDM\label{sec:sldm}}

\subsection{A motivating example\label{sec:example}}

Throughout the following sections we use a small example derived from \name{DBpedia} \cite{dbpedia}.
This example is about five books authored by J. R. R. Tolkien: \emph{The Hobbit}, \emph{The Silmarillion}, and three volumes of \emph{The Lord of the Rings}, namely \emph{The Fellowship of the Ring}, \emph{The Two Towers}, and \emph{The Return of the King}.
The whole graph we will use is presented in Listing \ref{lst:rdf-graph}.
This graph contains cherry-picked triples to clearly and concisely support the example.

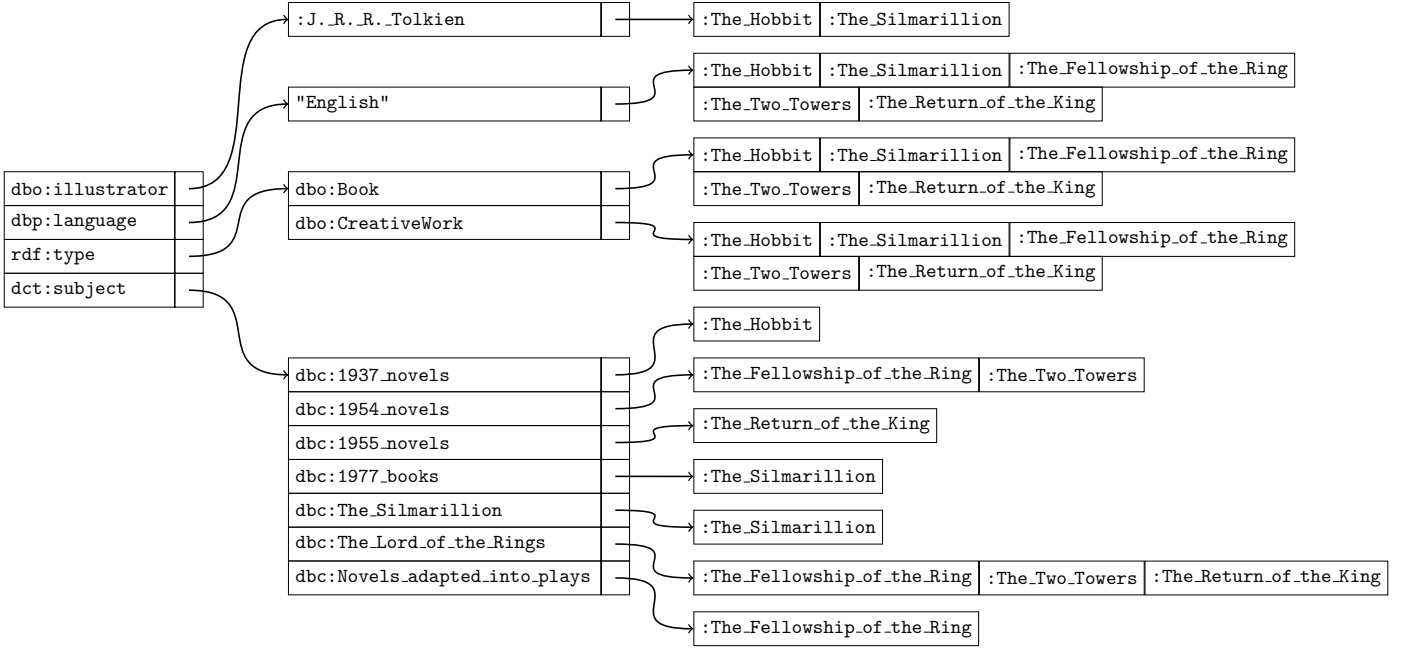
\begin{figure*}
\centering
\resizebox{\textwidth}{!}
{
\begin{tikzpicture}[yscale=.6,
subject/.style={minimum height=.6cm,rectangle,draw,node distance=0cm}
]
\foreach \x/\l in {0/dbo:illustrator,1/dbp:language,2/rdf:type,3/dct:subject}
{
	\draw (-1,-\x-6) rectangle ++(3,1) rectangle ++(.5,-1);	
	\node[anchor=west] at ($(-1,-\x-6)+(0,.5)$) {\texttt{\l}};
}
\foreach \dst [count=\src] in {0,2.5,5,10.5}
{
\draw[->,thick] ($(2.25,-\src+1-6+.5)$) .. controls ++(1.5,0) and ++(-1.5,0) .. ($(4,-\dst-.5)$);
}
\foreach \x/\l in {0/:J.\_R.\_R.\_Tolkien,2.5/"English",5/dbo:Book,6/dbo:CreativeWork,
10.5/dbc:1937\_novels,11.5/dbc:1954\_novels,12.5/dbc:1955\_novels,13.5/dbc:1977\_books,14.5/dbc:The\_Silmarillion,15.5/dbc:The\_Lord\_of\_the\_Rings,16.5/dbc:Novels\_adapted\_into\_plays}
{
	\draw (4,-\x-1) rectangle ++(5.5,1) rectangle ++(.5,-1);	
	\node[anchor=west] at ($(4,-\x-1)+(0,.5)$) {\texttt{\l}};
}
\subjects{illustrator-}{(11,-0.5)}{{\Hobbit,\Silmarillion}}
\draw[->,thick] ($(9.75,-1+.5)$) -- (illustrator-1.west);
\subjects{english-1-}{(11,-2)}{{\Hobbit,\Silmarillion,\FotR}}
\subjects{english-2-}{(11,-3)}{{\TT,\RotK}}
\draw[->,thick] ($(9.75,-3.5+.5)$) .. controls ++(1.5,0) and ++(-1.5,0) .. (english-1-1.west);
\subjects{book-1-}{(11,-4.5)}{{\Hobbit,\Silmarillion,\FotR}}
\subjects{book-2-}{(11,-5.5)}{{\TT,\RotK}}
\subjects{cw-1-}{(11,-7)}{{\Hobbit,\Silmarillion,\FotR}}
\subjects{cw-2-}{(11,-8)}{{\TT,\RotK}}
\foreach \dst [count=\src] in {book,cw}
{
\draw[->,thick] ($(9.75,-5+.5-\src)$) .. controls ++(1.5,0) and ++(-1.5,0) .. (\dst-1-1.west);
}
\subjects{1937-}{(11,-9.5)}{{\Hobbit}}
\subjects{1954-}{(11,-11)}{{\FotR,\TT}}
\subjects{1955-}{(11,-12.5)}{{\RotK}}
\subjects{1977-}{(11,-14)}{{\Silmarillion}}
\subjects{Sil-}{(11,-15.5)}{{\Silmarillion}}
\subjects{LotR-}{(11,-17)}{{\FotR,\TT,\RotK}}
\subjects{plays-}{(11,-18.5)}{{\FotR}}

\foreach \dst [count=\src] in {1937,1954,1955,1977,Sil,LotR,plays}
{
\draw[->,thick] ($(9.75,-10.5+.5-\src)$) .. controls ++(1.5,0) and ++(-1.5,0) .. (\dst-1.west);
}
\end{tikzpicture}
}
\caption{A tree-level index based on the triples from Listing \ref{lst:rdf-graph} having one of the following subjects: \Hobbit, \Silmarillion, \FotR, \TT, \RotK\label{fig:real_index1}}
\end{figure*}

Assume that a set of URIs $\uris_1$ consists of the following five URIs:
\Hobbit, \Silmarillion, \FotR, \TT,\\ \RotK.
In Figure \ref{fig:real_index1}, a three-level index is presented, built from the triples having subjects in the set $\uris_1$.

\begin{listing*}
\caption{A simple RDF graph used in the examples, written in Turtle syntax \cite{turtle}.\label{lst:rdf-graph}}
\begin{tabtt}
\small
@prefix : <http://dbpedia.org/resource/> .
@prefix dbp: <http://dbpedia.org/property/> .
@prefix dbo: <http://dbpedia.org/ontology/> .
@prefix dct: <http://purl.org/dc/terms/> .
@prefix dbc: <http://dbpedia.org/resource/Category:> .
@prefix skos: <http://www.w3.org/2004/02/skos/core#> . 
\setlength{\tabcolsep}{2pt}
\begin{tabular}{lll}
:The\_Hobbit & dbo:illustrator & :J._R._R._Tolkien ; \\
& dbp:language & "English" ; \\
& rdf:type & dbo:Book, dbo:CreativeWork ; \\
& dct:subject & dbc:1937_novels . \\
:The\_Fellowship\_of\_the\_Ring & dbp:language & "English" ; \\
& rdf:type & dbo:Book, dbo:CreativeWork ; \\
& dct:subject & dbc:1954_novels, dbc:The_Lord_of_the_Rings , \\
&& dbc:Novels_adapted_into_plays . \\
:The\_Two\_Towers & dbp:language & "English" ; \\
& rdf:type & dbo:Book, dbo:CreativeWork ; \\
& dct:subject & dbc:1954_novels , \\
& & dbc:The_Lord_of_the_Rings  . \\
:The\_Return\_of\_the\_King & dbp:language & "English" ; \\
& rdf:type & dbo:Book, dbo:CreativeWork ; \\
& dct:subject & dbc:1955_novels , \\
&& dbc:The_Lord_of_the_Rings  . \\
:The\_Silmarillion & dbo:illustrator & :J._R._R._Tolkien ; \\
& dbp:language & "English" ; \\
& rdf:type & dbo:Book, dbo:CreativeWork . \\
& dct:subject & dbc:1977_books , \\ 
& & dbc:The_Silmarillion . \\
dbc:1937_novels & rdf:type & skos:Concept . \\
dbc:1954_novels & rdf:type & skos:Concept . \\
dbc:1955_novels & rdf:type & skos:Concept . \\
dbc:1977_books & rdf:type & skos:Concept . \\
dbc:The_Silmarillion & rdf:type & skos:Concept . \\
dbc:The_Lord_of_the_Rings & rdf:type & skos:Concept . \\
dbc:Novels_adapted_into_plays & rdf:type & skos:Concept . \\
\end{tabular}
\end{tabtt}
\end{listing*}


\subsection{Frequent pattern\label{sec:frequent_pattern}}

\emph{Frequent pattern mining} is a task originating from data mining in databases.
A typical example considers a database of market baskets of items bought together.
The aim is to find all sets of items frequently bought together, which is occurring in at least given percent of the baskets in the database \cite{frequent_pattern_mining}.

In this work, we are interested in a slightly different approach.
We would like to find a set of possible superclasses for a given class, that is, we would like to find patterns partially describing a given set of objects in a given RDF graph.
By a \emph{pattern} we understand an arbitrary OWL 2 EL superclass expression $C$, as defined in Section \ref{sec:owl2el}.

To define a frequent pattern, we first define a \emph{matching function} $\match_\graph(a,C)$.
The intuitive meaning for the function is that $\match_\graph(a,C)=1$ iff an individual or a literal $a$ matches a pattern $C$ with respect to a graph $\graph$ and 0 otherwise.
The full definition of the function is presented in Table \ref{tab:match}.

\begin{table*}
\caption{The definition of the matching function $\match_\graph(a, C)$, where $a$ stands for an individual or a literal and $C$ is a pattern.\label{tab:match}}
\begin{minipage}{.49\textwidth}
\begin{align*}
\mu_\graph(a,DT) = & \begin{cases} 1 & a\in \text{value space of } DT \\ 0 & \text{otherwise} \end{cases} \\
\mu_\graph(a,LT[<= M]) = & \begin{cases} 1 & a\in \text{value space of } LT \land a \leq M \\ 0 & \text{otherwise} \end{cases} \\
\mu_\graph(a,GT[>= m]) = & \begin{cases} 1 & a\in \text{value space of } GT \land a \geq m \\ 0 & \text{otherwise} \end{cases}\\
    \mu_\graph(a,A)= & \begin{cases} 1 & (a,\type,A)\in\graph \\ 0 & \text{otherwise} \end{cases} \\
\mu_\graph(a,C\owlAnd D) = & \mu_\graph(a,C)\cdot\mu_\graph(a,D)
\end{align*}
\end{minipage}
\begin{minipage}{.49\textwidth}
\begin{align*}
\mu_\graph(a,\{b\})= & \begin{cases} 1 & a=b \\ 0 & \text{otherwise} \end{cases} \\
\mu_\graph(a,p\owlSome C) = & \begin{cases} 1 & \exists b\, (a,p,b)\in\graph\land \mu_\graph(b,C)=1 \\ 0 & \text{otherwise} \end{cases} \\
\mu_\graph(a,p\owlValue b) = & \begin{cases} 1 & (a,p,b)\in\graph \\ 0 & \text{otherwise} \end{cases} \\
\mu_\graph(a,p\owlSelf) = & \begin{cases} 1 & (a,p,a)\in\graph \\ 0 & \text{otherwise} \end{cases}
\end{align*}
\end{minipage}
\end{table*}

For example, consider the RDF graph presented in Listing \ref{lst:rdf-graph}.
\[\mu_\graph(\text{\texttt{:The\_Hobbit}, \texttt{dbo:Book}})=1\] because there is triple \texttt{(:The\_Hobbit, rdf:type, dbo:Book)} in the graph.
However, 
\[\begin{split}
\mu_\graph(\text{\texttt{:The\_Two\_Towers}, \texttt{dbo:illustrator}}\\\text{\owlValue \texttt{:J.\_R.\_R.\_Tolkien}})=0
\end{split}\] 
as there is no
 triple \texttt{(:The\_Two\_Towers, dbo:illustrator, :J.\_R.\_R.\_Tolkien)} in the graph.

We define frequency based on the measure of support.
Consider a set of URIs $\uris$ and a set of literals $\literals$.
A \emph{weighting function} $\weight$ is an arbitrary function $\weight\colon \uris\cup\literals\to\left[0, 1\right]$.
A \emph{support} of a subset $S\subseteq \uris\cup\literals$ given a weighting function $\weight$ is defined as:
\[ \support(S, \weight) = \sum_{s\in S}\weight(s) \]

Recall the graph from Listing \ref{lst:rdf-graph}.
Consider first a uniform weighting function $\weight_1$, such that $\weight_1(s)=\frac{1}{5}$ for all $s\in\uris_1$.
The three volumes of \emph{The Lord of the Rings} correspond to a set $S=\{$\FotR, \TT, \RotK$\}$, and its support is $\support(S, \weight)=\frac{3}{5}$.
The weighting function may be defined differently, for example,  let $\weight_2$ be $\frac{2}{6}$ for \FotR{} and $\frac{1}{6}$ for the rest of the set $\uris_1$.
In such a case, the support of the set $S$ would be $\support(S, \weight_2)=\frac{4}{6}$.
More details on the weighting function is given in Sections \ref{sec:core} and \ref{sec:initial_call}, where we show how it is used and recomputed in SLDM.

Observe, that given a set $\uris\cup\literals$ and a pattern $C$, the matching function defines the following set $S$:
\[ S = \{s\in\uris\cup\literals \colon \mu_\graph(s,C)=1\} \]
That is, $S$ is the subset of the set of URIs and literals $\uris\cup\literals$ such that all its elements  match the pattern $C$.
We will call such a set \emph{a proof set} for a pattern $C$.
This is to reflect the fact that $S$ proves that the corresponding pattern is frequent.
Then, the support of a pattern $C$ over the set $\uris\cup\literals$ can be defined as the following:
\[ \support(C, \weight) = \support(S, \weight) \]
A \emph{frequent pattern} $C$ over a set $\uris\cup\literals$ is any such  pattern that $\support(C, \weight)\geq \theta_\sigma$, where $\theta_\sigma$ is a minimal support threshold parameter.
A \emph{frequent predicate} $p$  over a set $\uris\cup\literals$ is any such a predicate that $\support(S,\weight)\geq \theta_\sigma$ for $S=\{s\in\uris\cup\literals\colon \exists o\, (s,p,o)\in\graph\}$.

Again, consider the graph from Listing \ref{lst:rdf-graph}, the set $\uris_1$ and the weighting function $\weight_2$, as defined above.
Let $C_1$ be a pattern \texttt{dbo:Book}.
Every of the five books occurs in a triple with the predicate \texttt{rdf:type} and the object \texttt{dbo:Book}, so the proof set $S_1=\uris_1$, and thus the support $\support(\texttt{dbo:Book}, \weight_2)=1$.
Now consider $C_2$ to be a pattern \texttt{dct:subject \owlValue dbc:1954\_novels}.
The proof set $S_2$ contains only \FotR{} and \TT and $\support(C_2, \weight_2)=\frac{2}{6}+\frac{1}{6}=\frac{3}{6}$.
If we assume that the minimal support threshold $\theta_\sigma$ is $0.8$, then the pattern $C_1$ \texttt{dbo:Book} is frequent, and the pattern $C_2$ \texttt{dct:subject \owlValue dbc:1954\_novels} is not frequent.
Following the same line of reasoning, we notice that the property \texttt{dbo:illustrator} has the support $\frac{2}{6}$ and is not frequent, whereas \texttt{dct:subject} is frequent with the support $1$.

By a depth $\len(C)$ of a pattern $C$ we understand the number of nested expressions in the pattern:
\begin{align*}
	\len(A) = & 1 \\
	\len(C\owlAnd D) = & \max\{\len(C),\len(D)\} \\
	\len(\{a\}) = & 1 \\
	\len(p\owlSome C) = & 1 + \len(C) \\
	\len(p\owlValue b) = & 1 \\
	\len(p\owlSelf) = & 1 \\
\end{align*}
Given a set of frequent patterns $\Patterns$, $C$ is the \emph{shallowest frequent pattern} in the set $\Patterns$ if and only if $\len(C)\leq\len(D)$ for all $D\in\Patterns$.
There may be multiple shallowest frequent patterns in the set $\Patterns$.
The set of frequent patterns $\Patterns$ can be infinite, so we mine only a finite subset of it defined by the shallowest frequent patterns.

\subsection{Language convention\label{sec:convention}}
In the following sections, we present a family of algorithms for mining the shallowest frequent patterns, as defined in the previous section.
These algorithms refer to a constant parameter $\theta_\sigma$, which is a minimal support threshold, as discussed in the previous section.

The keyword \texttt{yield} used in the pseudo-code snippets should be understood as \emph{add the argument to a temporary set and return this set at the end of the snippet}.
It is similar to the \texttt{yield} keyword in \emph{Python} programming language.

Every of the presented mining functions \texttt{Mine\ldots} and \SLDM function returns a pair or a set of pairs: a pattern and its proof set.
These proof sets are further used to combine patterns with $\owlAnd$, as it is explained in Section \ref{sec:conjunctions}.

In the examples, we use the set $\uris_1$ and the uniform weighting function $\weight_1$.

\subsection{Mining frequent datatype patterns\label{sec:datatype_patterns}}
We start the presentation of SLDM by presenting an algorithm for mining patterns $DT$, $LT[<=M]$, $GT[>=m]$.
This algorithm requires as an input a set of literals $\literals$ and a weighting function $\weight$.

\paragraph{Pattern $DT$}
For mining datatype patterns $DT$, we propose a solution based on the definition of lexical spaces of the datatypes \cite{xml_datatypes,owl2}.
For every untyped literal $l\in\literals$ we check with the grammar of every allowed datatype  if $l$ is a valid literal of this datatype.
This way, we generate a set $D_l$ of possible datatypes for every literal.
For a typed literal $l\in\literals$, the set $D_l$ contains only the specified type of $l$.
For a literal with a language tag, according to the specification \cite{Hawke:12:ADR}, the correct datatype is \texttt{rdf:PlainLiteral}, and so $D_l$ contains only this type.

\paragraph{Patterns $GT$, $LT$}
Having literals assigned to one or more datatype, we can mine patterns with relations $\geq$ and $\leq$.
For a subset $L_{GT}\subseteq \literals$ (resp. $L_{LT}\subseteq \literals$) containing all literal in $\literals$ of a type $GT$ (resp. $LT$), we look for a minimal (resp. maximal) value $m$ (resp. $M$) in the set $L_{GT}$ (resp. $L_{LT}$).
The type $GT$ (resp. $LT$) must be one of the allowed types defined in the Section \ref{sec:owl2el}.
A pattern $GT[>=m]$ (resp. $LT[<=M]$) by definition has support equal to the support of the set $L_{GT}$ (resp. $L_{LT}$).

The pseudo-code for mining $DT$, $LT$, and $GT$ patterns is presented in Algorithm \ref{alg:pattern_dt}.

\begin{algorithm}
\Fn{\MineDatatype{$\literals$, $\weight$}}
{
\ForEach{supported datatype $DT$}
{
$\support_{DT}\leftarrow 0$ \tcp*{support of the datatype}
$S_{DT}\leftarrow \emptyset$ \tcp*{proof set}
$m_{DT}\leftarrow \text{undefined}$ \tcp*{minimal value in the datatype}
$M_{DT}\leftarrow \text{undefined}$ \tcp*{maximal value in the datatype}
}
\ForEach{$l\in\literals$}
{
$D_l\leftarrow \emptyset$ \tcp*{datatypes for $l$}
\If{$l$ has type $t$}{$D_l\leftarrow \{t\}$}
\ElseIf{$l$ has language tag}{$D_l\leftarrow \{\text{\texttt{rdf:PlainLiteral}}\}$\;}
\Else{
\ForEach{supported datatype $DT$}
{
\If{$l$ matches the grammar of $DT$}{$D_l\leftarrow D_l\cup\{DT\}$}
}
}
\ForEach{$t\in D_l$}
{
$\support_t\leftarrow \support_t+\weight(l)$ \;
$S_{DT}\leftarrow S_{DT}\cup\{l\}$ \;
\If{$t$ can occur in $GT$ patterns and ($m_{DT}$ is undefined or $l<m_{DT}$)}{$m_{DT}\leftarrow l$}
\If{$t$ can occur in $LT$ patterns and ($M_{DT}$ is undefined or $M_{DT}<l$)}{$M_{DT}\leftarrow l$}
}
}
\ForEach{supported datatype $DT$}
{
\If{$\support_{DT}\geq\theta_\sigma$}{
\If{$m_{DT}$ and $M_{DT}$ are defined}
{
\tcp{It is impossible for $M_{DT}$ to be defined and $m_{DT}$ to be undefined at the same time, as every datatype that can occur in $LT$ patterns, can also occur in $GT$ patterns.}
\Yield{$DT \owlAnd GT[>=m_{DT}]\owlAnd LT[<=M_{DT}], S_{DT}$}
}
\ElseIf{$m_{DT}$ is defined}{\Yield{$DT \owlAnd GT[>=m_{DT}], S_{DT}$}}
\Else
{
\Yield{$DT, S_{DT}$}
}
}
}
}
\caption{The algorithm to find all frequent $DT$, $LT$ and $GT$ patterns given a set of literals $\literals$ and a weighting function $\weight$\label{alg:pattern_dt}
}
\end{algorithm}

\subsection{Mining frequent patterns for a given frequent predicate and index\label{sec:normal_patterns}}

In this section, we present four algorithms for mining frequent patterns for a given frequent predicate $p$ and with a fixed three-level index $I$.
Patterns that can be mined this way are: $A$, $p\owlValue b$, $r\owlValue l$, $p\owlSelf$, $\{b\}$, $\{l\}$.

\paragraph{Pattern $A$}
Following the defintion of the matching function $\mu$, a proof set for a pattern $A$ is defined as $S=\{s\in\uris\colon (s,\type,A)\in\graph\}$.
Finding all such frequent patterns can be accomplished with the algorithm presented in Algorithm~\ref{alg:pattern_A}.
This algorithm can be applied only if the frequent predicate $p$ is \type, otherwise the algorithm described in the next paragraph should be used.

Recall the index from Figure \ref{fig:real_index1} and consider $p$ to be the predicate \type.
There are two values in the second level: \texttt{dbo:Book} and \texttt{dbo:CreativeWork}.
Summing the weights of the URIs in the third level, we obtain 1 in both cases, and thus we mine patterns \texttt{dbo:Book} and \texttt{dbo:CreativeWork}.

\begin{algorithm}
\Fn{\MineType{$\Index, \weight$}}
{
\ForEach{$A\in \Index[\text{\type}]$}
{
	$\sigma \leftarrow 0$ \tcp*{a variable to compute support}
	\ForEach{$s\in \Index[\text{\type}][A]$}
	{
		$\sigma \leftarrow \sigma + \weight(s)$ \;
	}
	\If{$\sigma\geq\theta$}{\Yield{$A, \Index[\type][A]$}}
}
}
\caption{The algorithm to find all frequent patterns $A$, given a three-level index $\Index$ and a weighting function $\weight$.\label{alg:pattern_A}}
\end{algorithm}

\paragraph{Patterns $p \owlValue b$ and $r \owlValue l$} 
Following the definition of the matching function $\mu$, a proof set for a pattern $p \owlValue b$ is defined as $S=\{s\in\uris\colon (s,p,b)\in\graph\}$.
The very same reasoning applies to the patterns $r \owlValue l$.
Finding all such frequent patterns can be accomplished with the algorithm presented in Algorithm~\ref{alg:pattern_owl_value}.
This algorithm can be applied only if the frequent predicate $p$ is not \type.

Recall the index from Figure \ref{fig:real_index1} and let $p$ be the predicate \texttt{dbp:language}.
There is exactly one possible value in the second level, that is, \texttt{"English"}, and sum of the weights of the URIs in the third level of the index is 1.
Thus, we mine a pattern \texttt{dbp:language \owlValue "English"}.

Note that Algorithms \ref{alg:pattern_A} and \ref{alg:pattern_owl_value} differ only in the resulting patterns, not in the idea of operation.

\begin{algorithm}
\Fn{\MineValue{$\Index, p, \weight$}}
{
\ForEach{$b\in \Index[p]$}
{
	$\sigma \leftarrow 0$ \tcp*{variable to compute support}
	\ForEach{$s\in \Index[p][b]$}
	{
		$\sigma \leftarrow \sigma + \weight(s)$ \;
	}
	\If{$\sigma\geq\theta_\sigma$}{
	\Yield{$p\owlValue b, \Index[p][b]$}
	}
}
}
\caption{The algorithm to find all frequent patterns $p \owlValue b$, given a three-level index $\Index$, a frequent predicate $p$, and a weighting function $\weight$.\label{alg:pattern_owl_value}}
\end{algorithm}

\paragraph{Pattern $p\owlSelf$}
Following the definition of the matching function $\mu$, a proof set for a pattern $p \owlSelf$ is defined as $S=\{s\in\uris\colon (s,p,s)\in\graph\}$.
Finding all such frequent patterns can be accomplished with the algorithm presented in Algorithm~\ref{alg:pattern_owl_self}.

\begin{algorithm}
\Fn{\MineSelf{$\Index, p, \weight$}}
{
$\sigma \leftarrow 0$ \tcp*{variable to compute support}
$S\leftarrow \emptyset$ \tcp*{proof set}
\ForEach{$s\in \Index[p]$}
{
	\If{$s\in \Index[p][s]$}
	{
	$\sigma \leftarrow \sigma + \weight(s)$ \;
	$S\leftarrow S\cup \{s\}$ \;
	}
}
\If{$\sigma\geq\theta$}{\Return{$p\owlSelf$, S}}
}
\caption{The algorithm to find all frequent patterns $p \owlSelf$, given a three-level index $I$, a frequent predicate $p$ and a weighting function $\weight$.\label{alg:pattern_owl_self}}
\end{algorithm}

\paragraph{Patterns $\{b\}$ and $\{l\}$}
Following the definition of the matching function $\mu$, a proof set for a pattern $\{b\}$ (resp. $\{l\}$) is defined as $S=\{b\}\cap\uris$ (resp. $S=\{l\}\cap\literals$).
Finding all such frequent patterns can be accomplished with the algorithm presented in Algorithm~\ref{alg:pattern_b}.

\begin{algorithm}
\Fn{\MineEnum{$\uris$, $\literals$, $\weight$}}
{
\ForEach{$b\in \uris\cup\literals$}
{
	\If{$\weight(b)\geq\theta_\sigma$}{\Yield{$\{b\}, \{b\}$}}
}
}
\caption{The algorithm to find all frequent patterns $\{b\}$, given a set of URIs $\uris$, a set of literals $\literals$ and a weighting function $\weight$.\label{alg:pattern_b}}
\end{algorithm}

\subsection{Mining frequent conjunctions\label{sec:conjunctions}}

We build patterns that use conjunction, that is, $\owlAnd$, by combining patterns with identical proof sets mined earlier for given sets of URIs and literals.
It can be efficiently accomplished by using a hash map.
Following naming convention from frequent pattern mining, we call such a conjunction a \emph{closed conjunction}, because it cannot be extended (i.e., no other frequent pattern can be added to the conjunction) without shrinking the proof set and thus decreasing the support.
The corresponding algorithm is presented in Algorithm \ref{alg:mine_closed_conjunctions}.
As an input, the algorithm requires a set of frequent patterns and their proof sets $\Patterns$, and outputs all closed conjunctions that can be built from this set along with their proof sets.

In Section \ref{sec:normal_patterns}, we mined the following three patterns: \texttt{dbo:CreativeWork}, \texttt{dbo:Book} and \texttt{dbp:language \allowbreak\owlValue\allowbreak "English"}.
All of them have exactly the same proof set, consisting of all five books, so we can safely join them into a single pattern: \texttt{dbo:CreativeWork \owlAnd dbo:Book \owlAnd dbp:language\owlValue "English"}
\begin{algorithm}
\Fn{\MineClosedConjunctions{$\Patterns$}}
{
$G\leftarrow \emptyset$ \tcp*{a map from proof sets to patterns}
\ForEach{$P,S\in\Patterns$}
{
$G[S]\leftarrow G[S]\cup\{P\}$
}
\ForEach{$S\in G$}
{
$n\leftarrow \left|G[S]\right|$ \tcp*{a number of patterns in $G[S]$}
\Yield{$G[S][0]\owlAnd G[S][1]\owlAnd \ldots \owlAnd G[S][n], S$}
}
}
\caption{The algorithm finding all closed conjunctions from a set of patterns $\Patterns$.\label{alg:mine_closed_conjunctions}}
\end{algorithm}

\subsection{The core algorithm\label{sec:core}}

In Section \ref{sec:index}, we discussed how to build a three-level index for a given set of URIs.
In Sections \ref{sec:datatype_patterns}--\ref{sec:conjunctions}, we proposed a family of algorithms for mining frequent patterns, all except these using $\owlSome\!$.
Now, we combine these results into SLDM: we show how to remove some unnecessary information from the index, apply the mining algorithms defined earlier and use SLDM recursively to mine $\owlSome$ patterns.

The core of SLDM is presented in Algorithm \ref{alg:sldm}.
Its input consists of a set of URIs $\uris$, a set of literals $\literals$, and a weighting function $\weight$.
It outputs a set of shallowest frequent patterns for the set $\uris\cup\literals$.
To decide if a pattern is frequent, the algorithm also uses a minimal support threshold parameter $\theta_\sigma$.

To ensure that the algorithm ends, we add a maximal depth threshold parameter $\theta_\len$, and require that for all patterns $C$, the pattern depth $\len(C)$ does not exceed the parameter $\theta_\len$.
Recall that the pattern depth increases only with $\owlSome$ constructor, so this requirement is equivalent with limiting maximal recursion level.
To understand how a pattern of an infinite depth could be mined, consider an RDF graph with the following two triples: $\{(a,p,b), (b,q,a)\}$, an uniform weighting function $\weight(a)=\weight(b)=\frac{1}{2}$ and a minimal support threshold $\frac{1}{4}$.
Starting from $a$, SLDM follows to $b$ using the predicate $p$, then goes back from $b$ to $a$ using the predicate $q$ and again explores $a$, effectively getting stuck in an infinite loop while constructing a pattern $p \owlSome{} (q \owlSome{} (p\owlSome{} \ldots))$.
Observe that every time the minimal support threshold is exceeded there is no stopping condition if the maximal depth threshold is not considered.

The algorithm starts by mining patterns that does not require obtaining any triples from the SPARQL endpoint, as both \MineEnum and \MineDatatype do not use an index.
Then, it obtains new triples and builds a three-level index, following the idea described in Section \ref{sec:index}.
To store in the index only frequent predicates it uses a pruning function, that is presented in Algorithm \ref{alg:prune_index}.
For the index presented in Figure \ref{fig:real_index1}, it means that the predicate \texttt{dbo:illustrator} is removed, because sum of weights for \Hobbit{} and \Silmarillion{} does not exceed the threshold $\theta_\sigma$, and so the predicate is not frequent.

Later, SLDM iterates over each predicate in the index and uses functions defined in Section \ref{sec:normal_patterns}.
If for a given frequent predicate $p$ none of these functions yield a frequent pattern, the algorithm tries to mine deeper patterns with $\owlSome{}$.
Such an approach is to ensure that SLDM mines only the shallowest frequent patterns.
For this, a function \MineSome defined in Algorithm \ref{alg:pattern_some} is used.
Finally, obtained patterns are combined with $\owlAnd{}$ operator using \MineClosedConjunctions function, defined in Section \ref{sec:conjunctions}.

\begin{algorithm}
\Fn{\SLDM{$\uris$, $\literals$, $\weight$}}
{
$\Patterns\leftarrow$ \MineEnum{$\uris, \literals, \weight$} $\cup$ \MineDatatype($\literals$, $\weight$)  \;
\text{\small Obtain a set of triples $\triples$, such that their subjects are in $\uris$}\;
$\Index \leftarrow$ \BuildIndex{$\triples$} \;
$\Index \leftarrow$ \PruneIndex{$\Index, \weight$} \;
\ForEach(\tcp*[h]{Iterate over all frequent predicates}){$p\in\Index$}
{
$\Patterns_p\leftarrow\emptyset$ \;
\If{$p$ is \type}
{
$\Patterns_p\leftarrow$ \MineType{$\Index, \weight$}
}
\Else
{
$\Patterns_p\leftarrow $\MineValue{$\Index, p, \weight$} $\cup$ \MineSelf{$\Index, p, \weight$}
}
\If{$\Patterns_p\equiv\emptyset$ and the recursion level does not exceed $\theta_\len$}
{
$\Patterns_p\leftarrow$\MineSome{$\uris, \literals, \Index, \weight, p$} \;
}
$\Patterns\leftarrow\Patterns\cup\Patterns_p$\;
}
\Return{\MineClosedConjunctions{$\Patterns$}}
}
\caption{Swift Linked Data Miner algorithm.\label{alg:sldm}}
\end{algorithm}

\begin{algorithm}
\Fn{\PruneIndex{$\Index,\weight$}}
{ 
\ForEach{$p\in \Index$}
{
$S\leftarrow \emptyset$ \tcp*{A proof set for the predicate}
\ForEach{$o\in \Index[p]$}
{
    $S\leftarrow S\cup \Index[p][o]$ \;
}
$\sigma\leftarrow \emptyset$ \tcp*{A support for the predicate}
\ForEach{$s\in S$}
{
    $\sigma\leftarrow \sigma+\weight(s)$ \;
}
\If{$\sigma<\theta_\sigma$}
{
\tcp{$p$ is infrequent, we can drop the triples with it}
delete $\Index[p]$ \;
}
}
\Return $\Index$
}
\caption{The algorithm pruning a three-level index $\Index$ to store only frequent predicates according to a weighting function $\weight$.
The resulting index contains only frequent predicates.\label{alg:prune_index}
}
\end{algorithm}

\paragraph{Patterns $p\owlSome C$ and $r\owlSome DT$}
Finding such frequent patterns requires a recursive call to SLDM.
First, we must compute a new set of URIs $\uris_{new}$, a new set of literals $\literals_{new}$ and a new weighting function $\weight_{new}$.
Second, the recursive call is made, using $\uris_{new}$, $\literals_{new}$, and $\weight_{new}$ as the input.
Finally, after the recursive call returns, the generated frequent patterns are prefixed with $p\owlSome$, to obtain valid frequent patterns in current recursion level.
The corresponding algorithm is presented in Algorithm \ref{alg:pattern_some}.

\begin{algorithm}
\Fn{\MineSome{$\uris, \literals, \Index, p, \weight$}}
{
$\uris_{new}\leftarrow \emptyset$ \;
$\literals_{new}\leftarrow \emptyset$ \;
\ForEach{$n\in\uris\cup\literals$}{$den[n]\leftarrow 0$}
\tcp{First, we compute all denominators}
\ForEach{$o\in \Index[p]$}
{
\If{$o$ is an URI}
{$\uris_{new}\leftarrow \uris_{new}\cup \{o\}$}
\Else
{$\literals_{new}\leftarrow \literals_{new}\cup \{o\}$}
\ForEach{$s\in \Index[p][o]$}{$den[s] \leftarrow den[s] + 1$}
}
\tcp{Then, the actual new weights $\weight_{new}$ are computed}
\ForEach{$o\in \Index[p]$}
{
$\weight_{new}[o]\leftarrow 0$ \tcp*{we represent $\weight_{new}$ as a map}
\ForEach{$s\in \Index[p][o]$}{$\weight_{new}[o] \leftarrow \weight_{new}[o] + \frac{\weight(s)}{den[s]}$}
}
\tcp{Recursive call to the SLDM}
$\mathcal{P}\leftarrow$ SLDM($\uris_{new}, \literals_{new}, \weight_{new}$) \;
\tcp{Final step, prefixing the patterns from recursion}
\ForEach{$C, S\in\mathcal{P}$}
{
$S'\leftarrow\emptyset$ \;
\ForEach{$s\in S$}
{
$S'\leftarrow S'\cup \Index[p][s]$
}
\Yield{$p\owlSome C, S'$}
}
}
\caption{The algorithm finding all frequent patterns $p\owlSome$ given a set of URIs $\uris$, a set of literals $\literals$, a three-level index $I$, a frequent predicate $p$ and a weighting function $\weight$. \label{alg:pattern_some}}
\end{algorithm}

Observe that $p$ is a frequent predicate specified as a parameter to the \MineSome,
$\uris_{new}$ is the set of all URIs occurring in the second level of the index, and $\literals_{new}$ is the set of all literals occurring in the second level of the index.
To compute $\weight_{new}$ for a given object $o\in\uris_{new}\cup\literals_{new}$, we sum over all subjects $s$ such that triple $(s,p,o)$ is in the index, that is, we sum over $\Index[p][o]$.
Weight of every subject $s$ is equally divided over all objects $x$, such that a triple $(s,p,x)$ is in the index $\Index$.
\[w_{new}(o)=\sum_{s\in I[p][o]} \frac{w(s)}{\left|\{x\colon (s,p,x)\in\Index\}\right|} \]

Recall the index in Figure \ref{fig:real_index1} and let $p$ be \texttt{dct:subject}.
In this case, the set $\Patterns_p$ in the \SLDM function is empty, and thus function \MineSome is called.
Recall that we use the set $\uris_1$ and the weighting function $\weight_1$.
The set $\uris_{new}$ will consist of all values appearing in the second level of the index for the property \texttt{dct:subject}: 
\texttt{dbc:1937\_novels, dbc:1954\_novels, dbc:1955\_novels, dbc:1977\_books, dbc:The\_Silmarillion, dbc:The\_Lord\_of\_the\_Rings, dbc:Novels\_adapted\_into\_plays}.
We also need to compute new weights $\weight_{new}$ for these URIs.
We do so by equally dividing the previous weights.
Recall that \texttt{dbc:1937\_novels} occurs for exactly one URI from the original set, that is, \texttt{:The\_Hobbit}, and there are no other triples with \texttt{:The\_Hobbit} and \texttt{dct:subject}, so \texttt{dbc:1937\_novels} receives all the weight of \texttt{:The\_Hobbit}, that is, $\frac{1}{5}$.
Recall that \texttt{dbc:Novels\_adapted\_into\_plays} also occurs exactly once, but for an URI that has three possible values for \texttt{dct:subject}, and so, it receives weight $\frac{1}{3}\cdot\frac{1}{5}=\frac{1}{15}$.
Recall that \texttt{dbc:1954\_novels} occurs for two URIs: \texttt{:The\_Fellowship\_of\_the\_Ring} and \texttt{:The\_Two\_Towers}.
It is one of the three possible values for \texttt{:The\_Fellowship\_of\_the\_Ring}, and one of the two possible values for \texttt{:The\_Two\_Towers}, so \texttt{dbc:1954\_novels} is assigned weight $\frac{1}{3}\cdot\frac{1}{5}+\frac{1}{2}\cdot\frac{1}{5}=\frac{1}{6}$.
\texttt{dbc:1955\_novels}, \texttt{dbc:1977\_books} and \texttt{dbc:The\_Silmarillion} occur for exactly one URI each, and all three are one of two values for \texttt{dct:subject} for their respective URIs.
They both receive half of the respective weight, that is $\frac{1}{2}\cdot\frac{1}{5}=\frac{1}{10}$.
Finally, \texttt{dbc:The\_Lord\_of\_the\_Rings} occurs for three URIs and is one of two values for two of them, and one of three values for one of them, so the weight is $\frac{1}{3}\cdot\frac{1}{5}+\frac{1}{2}\cdot\frac{1}{5}+\frac{1}{2}\cdot\frac{1}{5}=\frac{4}{15}$.
The new weights are presented in Table \ref{tab:new-weights}.

\begin{table}
\caption{The weights computed for the recursive call in the example.\label{tab:new-weights}}
\begin{tabular}{lr}
URI & weight \\
\hline
\texttt{dbc:1937\_novels} & $\frac{6}{30}$ \\
\texttt{dbc:1954\_novels} & $\frac{5}{30}$ \\
\texttt{dbc:1955\_novels} & $\frac{3}{30}$ \\
\texttt{dbc:1977\_books} & $\frac{3}{30}$ \\
\texttt{dbc:The\_Silmarillion} & $\frac{3}{30}$ \\
\texttt{dbc:The\_Lord\_of\_the\_Rings} & $\frac{8}{30}$ \\
\texttt{dbc:Novels\_adapted\_into\_plays} & $\frac{2}{30}$
\end{tabular}
\end{table}

\begin{figure*}
\resizebox{\textwidth}{!}
{
\begin{tikzpicture}[yscale=.6,
subject/.style={minimum height=.6cm,rectangle,draw,node distance=0cm}
]
\draw (0,0) rectangle ++(2,1) rectangle ++(.5,-1);
\node[anchor=west] at (0,.5) {\type};
\draw (3.5,0) rectangle ++(3,1) rectangle ++(.5,-1);
\node[anchor=west] at (3.5,.5) {\texttt{skos:Concept}};
\subjects{concept-1-}{(8,.5)}{{dbc:1937\_novels, dbc:1954\_novels, dbc:1955\_novels, dbc:1977\_novels}}
\subjects{concept-2-}{(8,-.5)}{{dbc:The\_Silmarillion, dbc:The\_Lord\_of\_the\_Rings, dbc:Novels\_adapted\_into\_plays}}
\draw[->,thick] (2.25,.5) -- (3.5,.5);
\draw[->,thick] (6.75,.5) -- (concept-1-1.west);
\end{tikzpicture}
}
\caption{An index generated in the recursive call of the algorithm in the example.\label{fig:real_index2}}
\end{figure*}
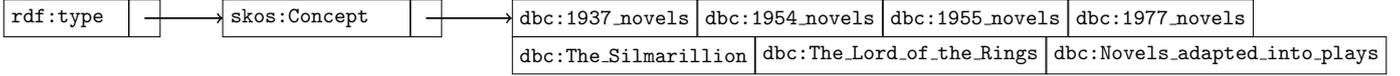

Now, we perform the recursive call.
We retrieve a new set of triples, having as subjects the URIs listed in Table \ref{tab:new-weights}.
The corresponding three-level index is presented in Figure \ref{fig:real_index2}.
Running the algorithm, we obtain a pattern \texttt{skos:Concept}.
After returning from the recursion, we prefix it and obtain a pattern \texttt{dct:subject} \owlSome \texttt{skos:Concept}.

\subsection{Initial call to the algorithm\label{sec:initial_call}}

In the initial call to the SLDM made by the user, we assume a simpler input, consisting only of a set of URIs $\uris$.
We assume an empty set of literals, because the goal of the user is to find patterns for given URIs  described by an RDF graph.
We also assume that a weighting function is uniform.
Finally, if the user already has some ontology $\mathcal{O}$, it may be desirable to remove the axioms that are already are entailed by the ontology.
These assumptions are combined and presented in Algorithm \ref{alg:sldm-call}.

\begin{algorithm}
\KwData{$\uris$ a set of URIs}
\KwResult{a set of frequent patterns}
$\weight\leftarrow \emptyset$ \tcp*{a map representing an initial weighting function}
$n\leftarrow \left|\uris\right|$ \;
\ForEach{$s\in\uris$}
{
$\weight[s]\leftarrow \frac{1}{n}$
}
\ForEach{$C,S\in\SLDM{$\uris, \emptyset, \weight$}$}
{
\If{$C$ is not entailed by $\mathcal{O}$}{\Yield{$C$}}
}
\caption{The suggested way of calling the SLDM algorithm by the user.\label{alg:sldm-call}}
\end{algorithm}

\section{Complexity analysis of the algorithm\label{sec:complexity}}
When computing the worst-case memory complexity  two factors that must be taken into account are as follows: the size of the indexes and the size of the sets of the mined patterns along with the corresponding proof sets.
Denote by $n$ the number of triples present in the RDF graph.
At any given point of the execution of the algorithm, the number of the recursive calls of SLDM is not greater than $\theta_\len$, that is, there are at most $\theta_\len$ three-level indexes in the memory.
Each of the indexes contain at most $n$ triples, thus the overall memory complexity of the indexes is $O(n^{\theta_\len})$.
For each triple present in the index, SLDM generates at most one pattern; thus, the number of patterns generated  is no greater than $n$.
The size of each proof set is bounded by the number of distinct URIs in the set $\uris$, which is also no greater than $n$.
Thus, in the worst-case, we arrive at $n$ patterns, having the support set of $n$ URIs each and the overall memory complexity of the mined patterns is $O(n^2)$.
The overall memory complexity of SLDM is $O(n^{\max\{\theta_\len, 2\}})$.

The reasoning behind the worst-case time complexity is similar: SLDM iterates over a three-level index, which is at most size $n$.
The sets $\uris$ and $\literals$ are also always at most of size $n$, and the access to hash-based maps is $O(1)$, so the upper bound on the overall time complexity of a single call to SLDM is $O(n)$.
Again, we must consider the recursive calls up to $\theta_\len$ deep, each of them able to perform up to $n$ recursive calls.
This way, we obtain a geometric series $n+n^2+\ldots+n^{\theta_\len}$ and conclude that the upper bound for SLDM worst-case time complexity is $O(n^{\theta_\len})$.
This analysis does not include the polynomial complexity of checking if an axioms is already entailed by the ontology.

\section{Relation to RDF Data Shapes\label{sec:shapes}}

There are applications where using OWL ontologies is not desirable or sufficient.
For example, one may want to avoid the complexity of reasoning procedures and only verify if a given RDF graph conforms to a model.
To address such use cases, RDF Data Shapes were proposed \cite{shacl_usecases}.
Their formal representation can be expressed in Shapes Constraint Language (\shacl{}) \cite{shacl}.
Throughout this section we use prefix \texttt{sh:} for the namespace \url{http://www.w3.org/ns/shacl#}  \cite{shacl}.

An OWL 2 EL superclass expression generated by SLDM can be transformed to \shacl{}.
Recall that SLDM results are based on online analysis of an RDF graph, so such a transformation is sanctioned.
We start by transforming the examples used in the previous section, and later, we give a full definition of the transformation.

\subsection{Example}

In Section \ref{sec:normal_patterns}, we mined three patterns: 
\texttt{dbo:CreativeWork}, \texttt{dbo:Book}, and \texttt{dbp:language \allowbreak\owlValue\allowbreak "English"}.
Each of them was mined because there were triples with a fixed predicate and a fixed object in the RDF graph, so it is natural to use this knowledge to map the patterns to the RDF shapes.
The obtained shapes are presented in Listing \ref{lst:shapes_normal}.
\begin{listing}
\caption{RDF Data Shapes expressed in \shacl, that correspond to the patterns \texttt{dbo:CreativeWork}, \texttt{dbo:Book} and \texttt{dbp:language\owlValue "English"} mined in Section \ref{sec:normal_patterns}.\label{lst:shapes_normal}}
\begin{tabtt}
:shape1 \type sh:Shape ;
        sh:property [
                sh:predicate rdf:type ;
                sh:hasValue dbo:CreativeWork ;
        ] .

:shape2 \type sh:Shape ;
        sh:property [
                sh:predicate rdf:type ;
                sh:hasValue dbo:Book ;
        ] .

:shape3 \type sh:Shape ;
        sh:property [
                sh:predicate dbp:language ;
                sh:hasValue "English" ;
        ] .
\end{tabtt}
\end{listing}

In Section \ref{sec:conjunctions}, we joined the abovementioned patterns with \owlAnd operator, obtaining the pattern
\texttt{dbo:CreativeWork \owlAnd dbo:Book \owlAnd dbp:language \allowbreak\owlValue\allowbreak "English"}.
The corresponding shape is presented in Listing \ref{lst:shapes_and}.

Finally, in Section \ref{sec:core}, we mined the pattern \texttt{dct:subject \owlSome skos:Concept}.
This was performed by first mining the pattern \texttt{skos:Concept} and then adding the prefix to it.
Therefore, we should follow the same principle during the transformation and refer to some other shape in the shape corresponding to the pattern \texttt{dct:subject \owlSome skos:Concept}.
The effect of the transformation is presented in Listing \ref{lst:shapes_some}.

\begin{listing}
\caption{An RDF Data Shape expressed in \shacl, that corresponds to the pattern \texttt{dbo:CreativeWork \owlAnd dbo:Book \owlAnd dbp:language\owlValue "English"} mined in Section \ref{sec:conjunctions}.
The shapes \texttt{:shape1, :shape2, :shape3} are defined in Listing \ref{lst:shapes_normal}.\label{lst:shapes_and}}
\begin{tabtt}
:shape4 \type  sh:Shape ;
        sh:constraint [
                sh:and(:shape1 :shape2 :shape3)
        ] .
\end{tabtt}
\end{listing}

\begin{listing}
\caption{An RDF Data Shape expressed in \shacl, corresponding to the pattern \texttt{dct:subject \owlSome skos:Concept} mined in Section \ref{sec:core}.
\label{lst:shapes_some}}
\begin{tabtt}
:shape5 \type sh:Shape ;
        sh:property [
                sh:predicate dct:subject ;
                sh:qualifiedValueShape [
                    sh:property [
                        sh:predicate rdf:type ;
                        sh:hasValue skos:Concept ;
                ]] ;
                sh:qualifiedMinCount 1 ;
        ] .
\end{tabtt}
\end{listing}

\subsection{Transformation}

Recall the definition of \emph{OWL 2 EL superclass expression $C$} from Section \ref{sec:owl2el}.
This is the grammar, which all patterns mined by SLDM follow.
Below, we define a recursive transformation from $C$ to \shacl.
This transformation can be done in post-processing, that is, first SLDM is used as defined in Section \ref{sec:sldm} or it can be incorporated  into the algorithms, changing them into algorithms directly mining RDF Data Shapes.
The transformation is defined as two functions: $\transform$ and $\transform_d$, that gets respectively a class expression $C$ or a data range $R$ as an argument and returns the corresponding \shacl{} expression.
The functions are fully defined in Table \ref{tab:shacl_transformation}, and  we give an explanation of this definition below.

We start by considering Algorithm \ref{alg:pattern_A}.
The algorithm yields the pattern $A$ any time when there is a large enough number of triples of form $(\cdot, \texttt{rdf:type}, A)$ in the graph.
The corresponding RDF Data Shape must take  all three properties into account: (a) a single triple, (b) with \texttt{rdf:type} as the predicate, and (c) with $A$ as the object.
We may thus define the function $\tau$ such that it yields such an expression every time its argument is a named class $A$, as specified in the first row of Table \ref{tab:shacl_transformation}.

Algorithm \ref{alg:pattern_owl_value} works in a very similar fashion, looking for a large number of triples of form $(\cdot, p, b)$ for fixed values of $p$ and $b$ and yielding the pattern $p\owlValue{} b$ in case of success.
The corresponding RDF Data Shape must look for a very similar property as before: a single triple, with $p$ as the predicate and $b$ as the object.
The corresponding \shacl{} expressions are specified as rows 2 and 3 in Table \ref{tab:shacl_transformation}.

Algorithm \ref{alg:pattern_owl_self} operates on a similar basis as the previous two, and counts the number of triples matching the pattern $(s, p, s)$ for a fixed value of $p$.
To the best of our knowledge, such a pattern cannot be expressed using core constraints defined by Knublauch and Ryman \cite{shacl}.
Fortunately, \shacl{} is, by design, extensible with \sparql{} queries.
The query must return RDF nodes that violate the constrain, that is, nodes $s$ such that a triple $(s, p, s)$ is not present in the graph.
The desired behavior can be achieved using SPARQL \texttt{sameTerm} function and FILTER NOT EXISTS clause.
The complete query, which returns a single binding for each triple violating the constraint, written in accordance with the requirements of \shacl{} specification and wrapped in appropriate RDF expressions, is presented in row 6 of Table \ref{tab:shacl_transformation}.

Algorithm \ref{alg:mine_closed_conjunctions} does not operate on triples directly, but rather combines the results of the recursive calls of SLDM to construct conjunctions.
We can follow the  same reasoning and  recursively transform the operands of the conjunction to \shacl{} and then combine them with \shacl{} \texttt{sh:and} operator.
For an OWL 2 EL class expression $C_1 \owlAnd{} C_2 \owlAnd{} \ldots$, each of the operands $C_1, C_2, \ldots$ must be first translated to \shacl{} using the $\tau$ function: $\transform(C_1), \transform(C_2), \ldots$ and only then they may be combined into a complex \shacl{} expression
specified as rows 4 and 10 in Table \ref{tab:shacl_transformation}.

Algorithm \ref{alg:pattern_b} also does not operate on triples, but only on the weighting function.
Every time it yields a pattern $\{a\}$ (resp. $\{l\}$) it means that a given URI $a$ (resp. a literal $l$) is frequent with respect to the function.
The corresponding SHACL expression must state that this URI (resp. literal) is expected to occur, and this can be achieved using SHACL \texttt{sh:in} operator, as stated in rows 5 and 11 of Table \ref{tab:shacl_transformation}.

The existential quantification cannot be directly expressed in SHACL; however, such an expression is semantically equivalent to a minimal cardinality restriction with caridinality 1 (for the proof, compare the semantics of \emph{ObjectMinCardinality} for $n=1$ and \emph{ObjectSomeValuesFrom}  \cite{owl2sem}).
Algorithm \ref{alg:pattern_some} uses recursion to first discover frequent patterns in a new subset of URIs or literals and then prefixes them with $p\owlSome$.
The same principle can be applied during the conversion to SHACL: first, the nested pattern must be converted using an appropriate function: $\transform$ in case of a class expression $C$ or $\transform_d$ in case of a data range $R$, obtaining an RDF node $\transform(C)$ or, respectively, $\transform_D(R)$.
Then, the result is wrapped in a SHACL expression specifying the predicate $p$ (using \texttt{sh:predicate}) and the minimal cardinality of 1 (\texttt{sh:qualifiedMinCount 1}).
The resulting transformation is presented in rows 7 (for object properties) and 8 (for data properties) of Table \ref{tab:shacl_transformation}. 

Finally, Algorithm \ref{alg:pattern_dt} counts triples having a literal of a given datatype $DT$ and finds appropriate minimal and maximal values.
The resulting RDF Data Shapes should thus verify if the value in question is of a given datatype and if it fits between the minimal and maximal value.
The expected datatype of a value can be expressed in \shacl{} using \texttt{sh:datatype} and the minimal and maximal value, using, respectively, \texttt{sh:minInclusive} and \texttt{sh:maxInclusive}.
The corresponding \shacl{} expressions are presented as rows 9, 12, and 13 of Table \ref{tab:shacl_transformation}.

\begin{table*}
\caption{A formal definition of functions $\transform$ and $\transform_d$, that is, a transformation from a SLDM pattern expressed as OWL 2 EL class expression to an RDF Data Shape expressed in \shacl{}.
Each of the top nodes in the transformation should be defined to be a \texttt{sh:Shape}, that is, there should be an additional triple \texttt{\ldots{} \type{} sh:Shape}.
We omitted them to make the table more readable.
\label{tab:shacl_transformation}}
\begin{tabular}{rp{.08\textwidth}p{.83\textwidth}}
\# & $C$ & $\transform(C)$ \\
\hline
1 & $A$ & \texttt{[sh:property [sh:predicate rdf:type; sh:hasValue $A$ ]]}\\
2 & $p \owlValue a$ & \texttt{[sh:property [sh:predicate $p$; sh:hasValue $a$;]]} \\
3 & $r \owlValue l$ & \texttt{[sh:property [sh:predicate $r$; sh:hasValue $l$;]]} \\
4 & $C_1\owlAnd C_2$ & \texttt{[sh:constraint [sh:and($\transform(C_1)\ \transform(C_2)$) ]]} \\
5 & $\{a\}$ & \texttt{[sh:constraint [sh:in ($a$)]]} \\
6 & $p \owlSelf$ & \texttt{[sh:constraint [\type{} sh:SPARQLConstraint;
                sh:sparql "SELECT \$this (\$this AS ?subject) ($p$ AS ?predicate) (?value AS ?object)
                WHERE \{
                        \$this $p$ ?value .
                        FILTER NOT EXISTS (sameTerm(\$this, ?value))
                \}"
        ]]} \\
7 & $p \owlSome C$ & \texttt{[sh:property [sh:predicate $p$; sh:qualifiedMinCount 1; sh:qualifiedValueShape $\transform(C)$ ]]} \\
8 & $r \owlSome R$ & \texttt{[sh:property [sh:predicate $p$; sh:qualifiedMinCount 1; sh:qualifiedValueShape $\transform_d(R)$ ]]} \\
\hline
\hline
\# & $R$ & $\transform_d(R)$ \\
\hline
9 & $DT$ & \texttt{[sh:constraint [sh:datatype $DT$;]]} \\
10 & $R_1\owlAnd R_2$ & \texttt{[sh:constraint [sh:and($\transform_d(R_1)\ \transform_d(R_2)$) ]]} \\
11 & $\{l\}$ & \texttt{[sh:constraint [sh:in ($l$)]]} \\
12 & $LT[<=l]$ & \texttt{[sh:constraint [sh:datatype $LT$; sh:minInclusive $l$]]} \\
13 & $GT[>=l]$ & \texttt{[sh:constraint [sh:datatype $GT$; sh:maxInclusive $l$]]} \\
\hline
\end{tabular}
\end{table*}

\section{\protege{} plugin\label{sec:protege}}

\begin{figure*}
\centering
\includegraphics[width=.9\textwidth]{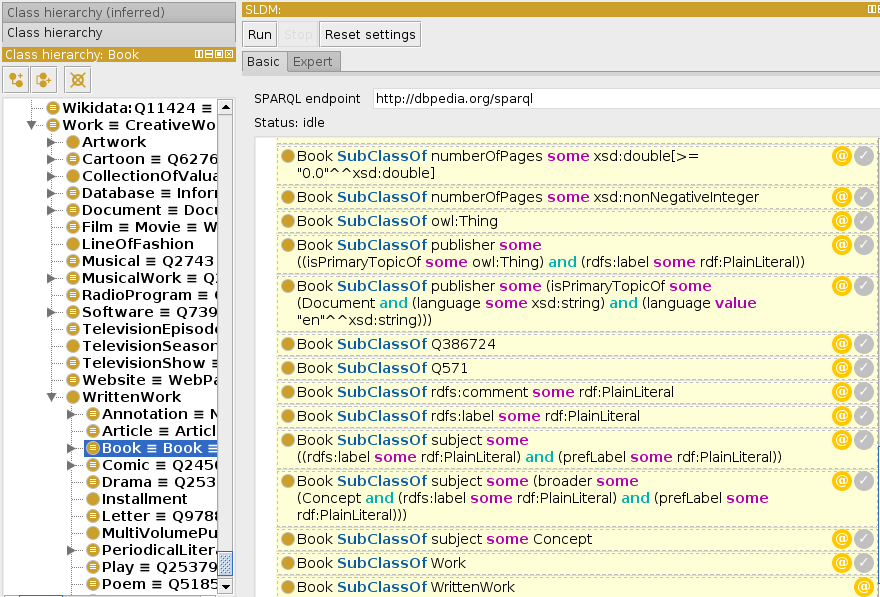}
\caption{The \protege{} plugin with some axioms mined for the class \texttt{dbo:Book} using the \name{DBpedia} endpoint.
On the right-hand side of a mined axiom there are two buttons: the first one (with \texttt{@} symbol) to display additional information about the axiom, for example, support, and the other one (with \cmark symbol) to add the axiom to the ontology.
The last axiom in the list has only the first button, because it already is asserted in the ontology.
\label{fig:protege_plugin}
}
\end{figure*}

\begin{figure*}
\centering
\includegraphics[width=.75\textwidth]{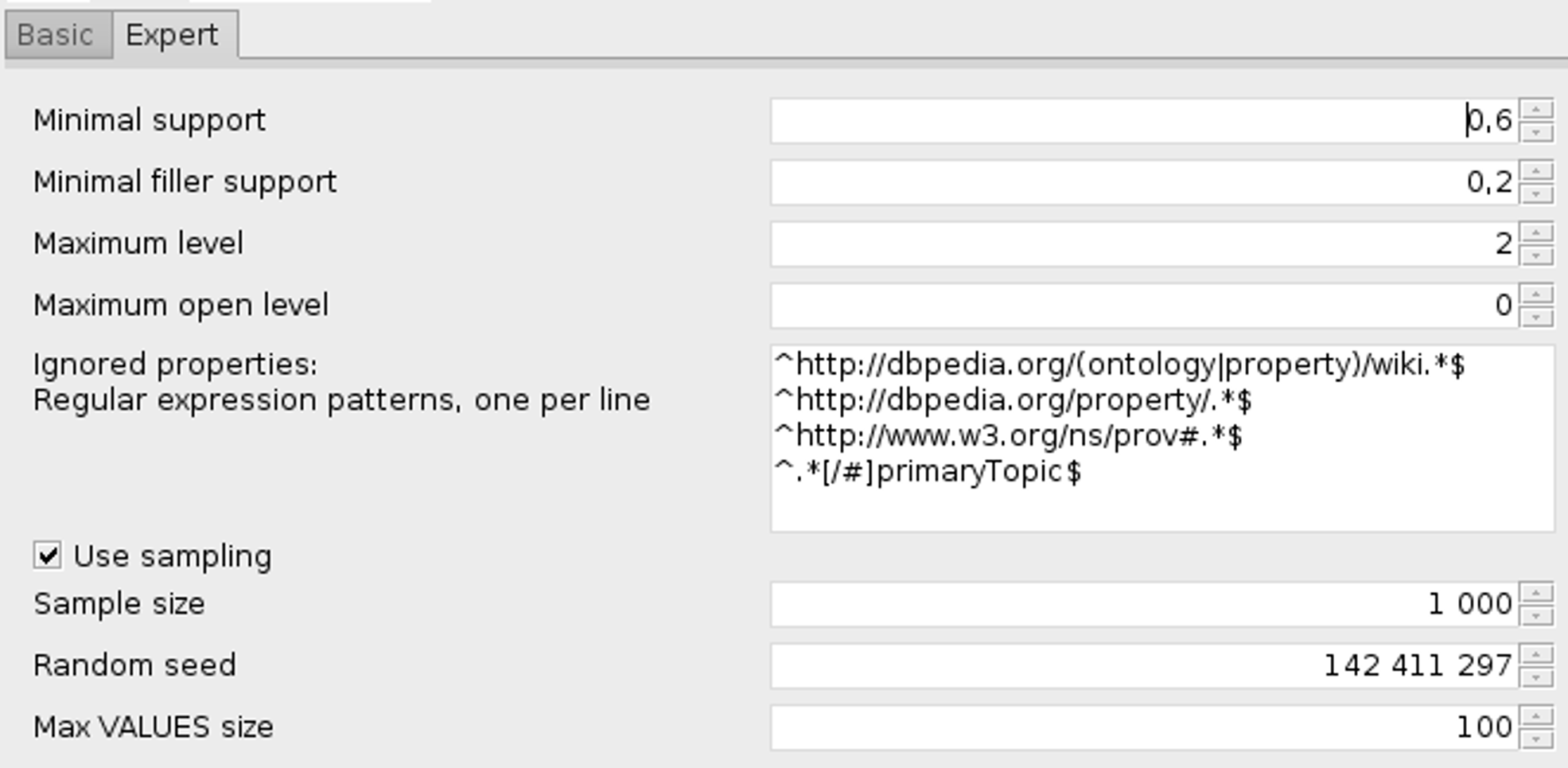}
\caption{The \texttt{Expert} tab of the \protege{} plugin.
The user can configure here a minimal support threshold, a maximal depth of the axioms, sampling parameters and predicates ignored during the mining (using regular expressions).\label{fig:expert}}
\end{figure*}

In Figure \ref{fig:protege_plugin}, we present a screenshot of the \protege{} plugin, which enables the user to mine new superclass expressions directly from \protege{}.
The user must just select a class in the class hierarchy view, enter the address of a \sparql{} endpoint she wants to use and click \texttt{Run}.
The axioms are displayed right after they are mined, not after the whole mining ends, so the user does not need to wait.
Mining can be stopped at any time by clicking \texttt{Stop} button.
On the \texttt{Expert} tab (Figure \ref{fig:expert}), it is possible to configure SLDM parameters, such as the minimal support threshold or if sampling is to be used.
One can also specify predicates that should be ignored during the mining.
For example, one may want to ignore provenance-related predicates (e.g., \texttt{dbo:wikiPageID}), as not directly related to the semantics of the mined class.

The plugin is written in \name{Java}, using the OWL API\footnote{\url{http://owlapi.sourceforge.net/}} \cite{owlapi} and \name{Apache Jena}\footnote{\url{https://jena.apache.org/}} \cite{jena}.
The source code of the plugin is available in \name{Git} repository \url{https://bitbucket.org/jpotoniec/sldm} along with the compilation and installation instructions.
There is also a link to a precompiled version of the plugin.




\section{Experimental evaluation\label{sec:experiment}}
\subsection{Extending the DBpedia ontology\label{sec:dbpedia}}

This section presents all the steps that were undertaken in order to prepare and conduct an experiment on a crowdsourcing platform \name{CrowdFlower}\footnote{\url{https://www.crowdflower.com/}}.
Our aim was to answer the following research question: can SLDM mine new, meaningful axioms, that can be added to the ontology.
To answer the question, we used \name{DBpedia} 2015-04 with the \name{DBpedia} ontology, and we followed the experimental protocol described:
\begin{enumerate}
\item We conducted exploratory data analysis to select a set of classes.
\item For the selected classes, we used SLDM to generate superclass expressions, and used them to obtain a set of \owlSubClassOf{} axioms for a selected class, with the class in the left-hand side, and an expression in the right-hand side of an axiom.
\item We translated the generated axioms into natural language sentences.
\item We generated test questions to ensure that participants of the experiment are paying attention to their tasks.
\item These sentences were then posed to \name{CrowdFlower} for verification by the contributors.
\item We collected and analyzed the results of the verification.
\end{enumerate}

In the following sections, the details of the experimental protocol are explained.

\subsubsection{Exploratory data analysis}
To select a set of classes from the \name{DBpedia} ontology, what would allow us to conduct a high quality, statistically reliable experimental evaluation, we performed exploratory data analysis.
For every class in the ontology, we computed the following characteristics using \name{DBpedia} 2015-04:
\begin{enumerate}
\item the number of class instances,
\item the number of different triples, for which the subject belongs to the class,
\item the number of different predicates, for which there exists a triple in the dataset with a given predicate, and the subject belonging to the class,
\item the depth of the class in the subsumption hierarchy in the \name{DBpedia} ontology (the shortest path from the root of the hierarchy \texttt{owl:Thing} to the class).
\end{enumerate}
Histograms of the obtained values are presented in Figures \ref{fig:hist1}--\ref{fig:hist4}.
On the basis of the histograms, we chose a set of criteria that the selected classes should fulfill.

To provide enough statistical support, we chose classes with more that 1000 instances and every instance occurring as a subject on average in more than 50 triples.
To avoid generating a very large number of axioms, which would increase the costs of the crowdsourcing verification, we decided to keep the number of different predicates in range from 20 to 35.
Finally,
we decided on selecting classes with the depth of at least 3, and in such a manner that all selected classes should have pairwise different parents and at least three different grandparents.

\begin{figure}
\centering
\tikzstyle{every node}=[draw=black,thick,anchor=west]

\begin{tikzpicture}[%
  grow via three points={one child at (0.5,-0.7) and
  two children at (0.5,-0.7) and (0.5,-1.4)},
  edge from parent path={(\tikzparentnode.south) |- (\tikzchildnode.west)}]
  \node[dashed] {\texttt{owl:Thing}}
    child { node {\texttt{dbo:Agent}}
      child { node {\texttt{dbo:Person}}
        child {node {\texttt{dbo:Journalist}}}
        }}
    child [missing]{}
    child { node {\texttt{dbo:Work}}
      child { node {\texttt{dbo:Software}}
        child {node {\texttt{dbo:Programminglanguage}}}}
      child [missing] {}
      child { node {\texttt{dbo:WrittenWork}}
        child {node {\texttt{dbo:Book}}}}
        }
    child [missing] {}
    child [missing] {}
    child [missing] {}
    child [missing] {}
    child { node {\texttt{dbo:TopicalConcept}}
      child { node {\texttt{dbo:Genre}}
        child {node {\texttt{dbo:MusicGenre}}}}
        }
    child [missing] {}
    child { node {\texttt{dbo:Place}}
      child { node {\texttt{dbo:NaturalPlace}}
        child {node {\texttt{dbo:Crater}}}}
        };
\end{tikzpicture}
\caption{The set of classes from the \name{DBpedia} ontology used in the experiment. 
\texttt{owl:Thing} is depicted in the picture only for reference and was not used in the experiment.}
\label{fig:hierarchy1}
\end{figure}
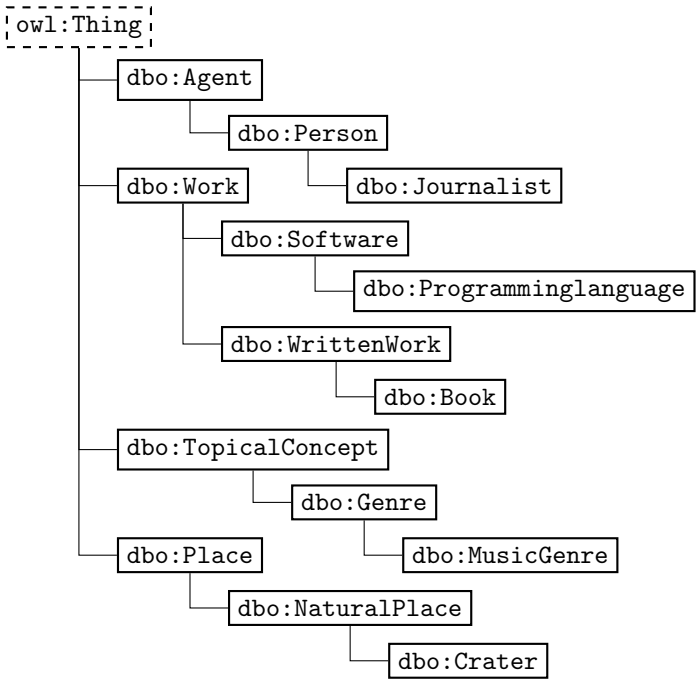

We selected 5 classes, which meet all the aforementioned criteria: \texttt{Journalist, ProgrammingLanguage, Book, MusicGenre, Crater} together with their ancestors: \texttt{Agent, Person, Work, Software, WrittenWork, TopicalConcept, Genre, Place, NaturalPlace}.
Full hierarchy is presented in Figure \ref{fig:hierarchy1}.
For each of these 14 classes, we used SLDM to generate two sets of axioms: the first one using the minimal support threshold $\theta_\sigma=0.5$ and the second one with  $\theta_\sigma=0.8$.
The obtained axioms are available in the \name{Git} repository \url{https://bitbucket.org/jpotoniec/sldm}, in the subfolder \texttt{CF\_source}.

To get some insights into the novelty of the mined axioms, we used HermiT reasoner\footnote{\url{http://www.hermit-reasoner.com/}} \cite{hermit} and for each of the mined set of axioms, we calculated how many of them were already logically entailed by the \emph{DBpedia} ontology and how many of them were logically entailed by the \emph{DBpedia} ontology enriched with the mined axioms for the superclass.
The detailed statistics are presented in Table \ref{tab:entailment_stats}.
For example, for the class Book and the minimal support threshold $\theta_\sigma=0.8$, 35 axioms were mined by SLDM, out of which 7 were logically entailed by the ontology and 15 were logically entailed by the ontology with the mined axioms for the class WrittenWork asserted.
It must be noted that in all the cases, SLDM was able to discover more than it was already present in the ontology.

\begin{figure*}
\begin{subfigure}[t]{.48\textwidth}
\centering
\begin{tikzpicture}[font=\small]
    \begin{axis}[
      ybar,
      bar width=20pt,
      xlabel={Number of instances},
      ylabel={Number of classes},
      ymin=0,
      ytick=\empty,
      xtick=data,
      axis x line=bottom,
      axis y line=left,
      every axis x label/.style={at={(axis description cs:0.5,-0.35)}},
      enlarge x limits=0.2,
      xticklabels={$0-10$,$10-10^2$,$10^2-10^3$,$10^3-10^4$,$10^4-10^5$,$10^5-10^6$, $10^6-10^7$},
      xticklabel style={align=center, rotate=90},
      nodes near coords={\pgfmathprintnumber\pgfplotspointmeta}
    ]
      \addplot[fill=white] coordinates {
        (0,15) (10, 40) (20, 128) (30, 168) (40, 78) 
        (50, 18) (60, 2)
      };
    \end{axis}
  \end{tikzpicture}
    \caption{A histogram of number of class instances in \name{DBpedia} for the classes from the \name{DBpedia} ontology.\label{fig:hist1}}
\end{subfigure}
\begin{subfigure}[t]{.48\textwidth}
\centering
\begin{tikzpicture}[font=\small]
    \begin{axis}[
      ybar,
      bar width=15pt,
      xlabel={Number of subjects},
      ylabel={Number of classes},
      ymin=0,
      ytick=\empty,
      xtick=data,
      axis x line=bottom,
      axis y line=left,
      every axis x label/.style={at={(axis description cs:0.5,-0.25)}},
      enlarge x limits=0.05,
      xticklabels={0--10,10--2,20--30,30--40,40--50, 50--60, 60--70, 70--80, 80--90, 90--100, 100--110},
      xticklabel style={align=center, rotate=90},
      nodes near coords={\pgfmathprintnumber\pgfplotspointmeta}
    ]
      \addplot[fill=white] coordinates {
        (0,6) (10, 10) (20, 74) (30, 175) (40, 104) 
        (50, 41) (60, 12) (70, 8) (80, 4) (90, 7) (100, 2)
      };
    \end{axis}
  \end{tikzpicture}
  \caption{A histogram of average number of different predicates, for which there is a triple in  \name{DBpedia} with the subject from a given class.\label{fig:hist2}}
\end{subfigure}
\begin{subfigure}[t]{.48\textwidth}
\centering
\begin{tikzpicture}[font=\small]
    \begin{axis}[
      ybar,
      bar width=15pt,
      xlabel={Number of triples},
      ylabel={Number of classes},
      ymin=0,
      ytick=\empty,
      xtick=data,
      axis x line=bottom,
      axis y line=left,
      every axis x label/.style={at={(axis description cs:0.5,-0.3)}},
      enlarge x limits=0.2,
      xticklabels={0--100, 100--200, 200--300, 300--400, 400--500, 500--600, 600-700, 700--800},
      xticklabel style={align=center, rotate=90},
      nodes near coords={\pgfmathprintnumber\pgfplotspointmeta}
    ]
      \addplot[fill=white] coordinates {
        (0,57) (10, 233) (20, 89) (30, 30) (40, 11) 
        (50, 12) (60, 8) (70, 4)
      };
    \end{axis}
  \end{tikzpicture}
\caption{
A histogram of average number of triples in \name{DBpedia} for which the subject belongs to a given class.\label{fig:hist3}}
\end{subfigure}
\begin{subfigure}[t]{.48\textwidth}
\centering
\begin{tikzpicture}[font=\small]
    \begin{axis}[
      ybar,
      bar width=20pt,
      xlabel={Level of the class},
      ylabel={Number of classes},
      ymin=0,
      ytick=\empty,
      xtick=data,
      axis x line=bottom,
      axis y line=left,
      every axis x label/.style={at={(axis description cs:0.5,-0.11)}},
      enlarge x limits=0.2,
      xticklabels={1,2,3,4,5,6},
      xticklabel style={align=center},
      nodes near coords={\pgfmathprintnumber\pgfplotspointmeta}
    ]
      \addplot[fill=white] coordinates {
        (0,32) (10, 82) (20, 106) (30, 188) (40, 36) 
        (50, 5)
      };
    \end{axis}
  \end{tikzpicture}
\caption{Level of classes in the subsumption hierarchy in the \name{DBpedia} ontology.}
\label{fig:hist4}
\end{subfigure}
\caption{Histograms for characteristics used to select the set of classes from the \name{DBpedia} ontology to perform the experiment on.}
\end{figure*}
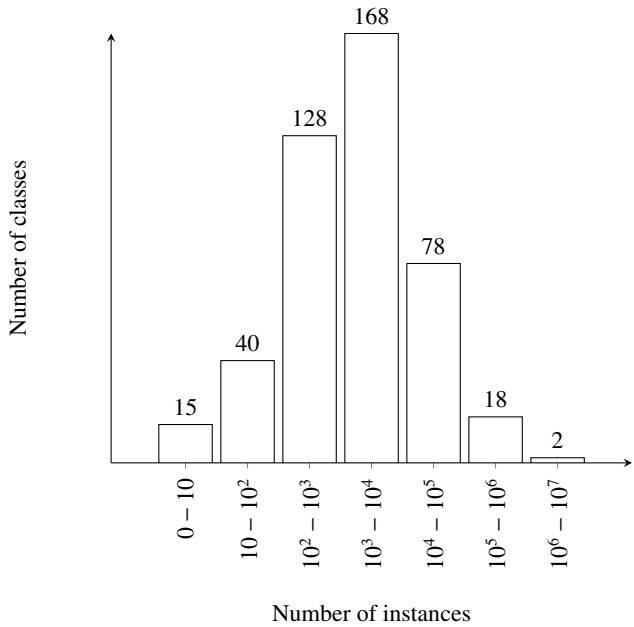
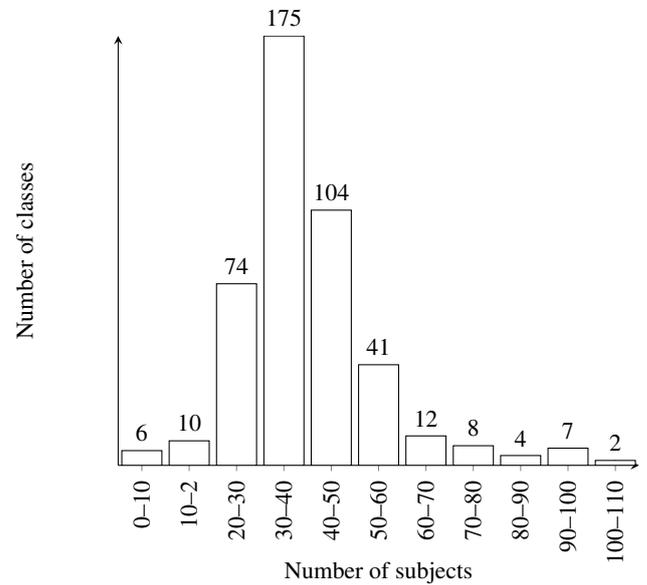
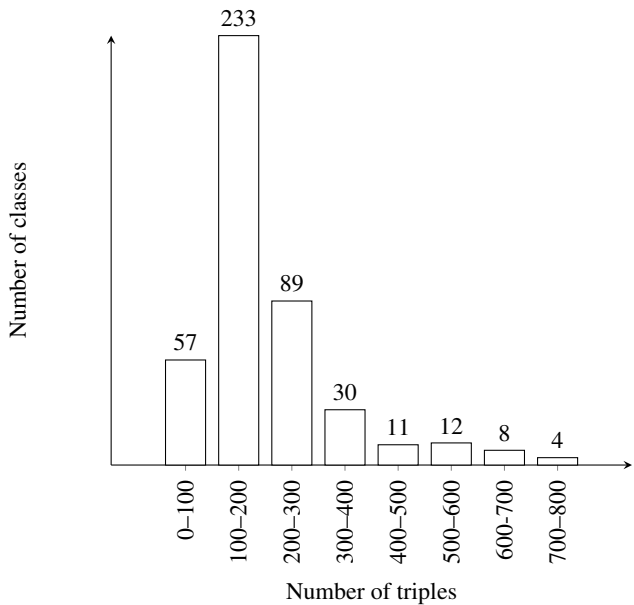
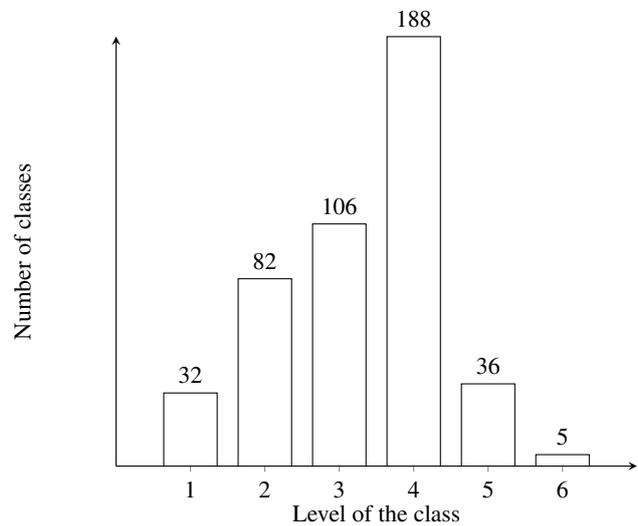

\begin{table*}
\caption{Number of axioms mined for each of the classes listed in Figure \ref{fig:hierarchy1} and each of the two minimal support thresholds $\theta_\sigma$.
The column \emph{overall} presents the overall number of mined axioms for a given class, the column \emph{in the ontology} presents how many of them were logically entailed by the \emph{DBpedia} ontology and, finally, the column \emph{in the superclass} presents the same, but also asserting all the mined axioms for the superclass.
}
\label{tab:entailment_stats}
\centering
\begin{tabular}{l|>{\centering\arraybackslash}p{1.4cm}>{\centering\arraybackslash}p{1.4cm}>{\centering\arraybackslash}p{1.4cm}|>{\centering\arraybackslash}p{1.4cm}>{\centering\arraybackslash}p{1.4cm}>{\centering\arraybackslash}p{1.4cm}}
Class & \multicolumn{3}{c|}{Number of mined axioms $\theta_\sigma=0.8$} & \multicolumn{3}{c}{Number of mined axioms $\theta_\sigma=0.5$} \\
& Overall & In the ontology & In the superclass & Overall & In the ontology & In the superclass \\
\hline
Agent & 9 & 3 & - & 22 & 3 & - \\
Person  & 9 & 7 & 9   & 22 & 7 & 20\\
Journalist  & 109 & 8 & 10 & 131 & 8 & 21 \\
\hline
Work  & 11 & 3 & - & 14 & 3 & - \\
Software & 17 & 4 & 12 & 45 & 4 & 14 \\
ProgrammingLanguage  & 16 & 6 & 12  & 28 & 6 & 13\\
WrittenWork  & 14 & 4 & 12 & 63 & 4 & 15\\
Book  & 35 & 7 & 15 & 108 & 7 & 29\\
\hline
TopicalConcept & 10 & 3 & - & 13 & 3 & -\\
Genre  & 13 & 4 & 11 & 27 & 4 & 13\\
MusicGenre  & 13 &  5 & 13  & 27 &  5 & 27\\
\hline
Place & 21 & 3 & - & 186 & 5 & -\\
NaturalPlace  & 13 &  4 & 12 & 24 &  4 & 24 \\
Crater & 12 & 5 & 11 & 25 & 5 & 12 \\
\hline
\end{tabular}
\end{table*}

\subsubsection{Translation of the axioms to natural language}
Ontological axioms expressed in OWL (e.g., using Turtle) cannot be easily understood by English speakers that are not familiar with the Semantic Web  technologies.
Therefore, we proposed a procedure of translation of OWL axioms to English.
We wanted to use a simple variant of the language, so we decided to choose Attempto Controlled English \citep{kaljurand2007verbalizing}.
It is a controlled version of normal English language that involves advantages of formal representation (well defined syntax, possibility of automatic processing) and natural language (expressiveness and ease of understanding) \citep{fuchs2008attempto}. 
Each person that knows basics of English should be able to understand a translated sentence without any knowledge about its formal representation.
Usage of controlled language has one other, very important feature: the translation is fully reversible, so we do not lose any information. 

We decided to create a whole translator on our own.
The reason for that was the fact that existing tools (e.g., \name{OWL Verbalizer}\footnote{\url{https://github.com/Kaljurand/owl-verbalizer}} \citep{owlverb}) are restricted and work only for some of our examples.

In the OWL axioms generated by SLDM, we identified a set of structural templates and for every template, we provided a corresponding template in English.
The URIs in the axioms were replaced by their corresponding labels during the translation.
The core idea of the translation tool is to analyze an axiom level by level and match it to the templates.
The output is a set of simple sentences, that represent more and more specific parts of constrains.
Sample axiom, that pertains \texttt{dbo:Journalist} class
\begin{alltt}
\texttt{dbo:Journalist} \owlSubClassOf{} \texttt{dbo:nationality} 
        \owlSome (\texttt{dbo:governmentType} \owlSome{} \texttt{owl:Thing}
        \owlAnd{} \texttt{dbo:leader} \owlSome \texttt{owl:Thing})
\end{alltt}
can be translated into sentences \textit{Every journalist has nationality. Nationality has government type. Nationality has leader.}
During the translation, we also performed some pruning  to make the final sentences more readable and limit the costs of the experiment.
We removed axioms that contained concepts that are characteristic for internal structure of \name{DBpedia} or act as metadata, for example, predicate \texttt{dbp:hasPhotoCollection}.
We reason that such axioms are very hard to understand for a non-expert, and thus cannot be efficiently verified by the contributors of a crowdsourcing platform.
We also removed axioms containing namespaces from other Linked Data sets, for example, \name{Wikidata} namespace, in order to decrease number of axioms to verify, and avoid displaying numerical URIs to the users, for example, we removed the axiom \texttt{dbo:Book} \owlSubClassOf{} \texttt{wikidata:Q1930187}\footnote{The prefix \texttt{wikidata:} corresponds to \url{http://www.wikidata.org/entity/}. 
The complete list of removed axioms is available in the \emph{Git} repository, in the file \texttt{CF\_source/removed\_axioms.txt}.
}.

After application of the translation tool to the axioms generated by SLDM, we obtained a set of sentences.
Each of these sentences was then used as a base to form a question.
A question consists of: a sentence, which is to be verified; a set of three allowed answers, from which only one is to be selected: \emph{Yes}, \emph{No}, \emph{I don't know}; an optional field to explain why a particular answer was selected.

\subsubsection{Test questions}
The quality of the results achieved from crowdsourcing experiment can be significantly improved by introducing test questions \citep{mortensen2013crowdsourcing}.
The right answer to these questions is known before the experiment.
They are used to check reliability of crowdsourcing platform contributors.
They can be used in two ways:
\begin{enumerate}
\item One prepares a quiz for the contributors, that contains only the test questions.
If a contributor passes the test, she is allowed to answer payable questions.
\item For each set of questions, that are presented to a contributor, one question is a test question.
If the contributor does not answer the test question correctly, the other answers from her are discarded, and she does not get paid for them.
\end{enumerate}
During the experiment, we used the second solution, because it requires contributor attention for every set of questions.
A good practice recommends having 10-20\% test questions in the input dataset\footnote{\url{http://www.success.crowdflower.com}}.
Some of our test questions were correct (i.e., required an answer \emph{Yes}) and some were incorrect (i.e., required an answer \emph{No}), in order to ensure that a contributor cannot select always the same answer and ignore the questions completely.

To obtain the test questions, we used reasoner  \emph{Pellet} \cite{pellet} to find in the set of axioms generated by SLDM axioms that logically follows from the ontology.
The questions corresponding to these axioms were then used as the test questions with a known correct answer \emph{Yes}.

To obtain test questions with a correct \emph{No} answer, we selected some of the axioms generated by SLDM containing only a named class in the right-hand side and replaced the class by some other, unrelated class, obtaining, for example, an axiom \texttt{dbo:Journalist} \owlSome{} \texttt{dbo:Book}.
We also generated some false axioms by adding \owl{not} to the left-hand side of an axiom inferred from the ontology, obtaining, for example, a sentence \emph{Not every software is software}.

\subsubsection{CrowdFlower experiment setup}

The last activity to do before starting a crowdsourcing experiment is to setup the settings of the experiment and create an instruction for the contributors.
Both of these steps are crucial with respect to ensuring quality of the experimental results.

We set up the settings in the following way:\footnote{See \url{https://success.crowdflower.com/hc/en-us/articles/201855719-Guide-to-Basic-Job-Settings-Page} for additional explanation of the settings}
As our questions are quite simple, we requested for contributors of the lowest level, as this allowed us to obtain the results faster.
We decided on presenting 10 questions at once (i.e., on a single page) to a single contributor, as it should not take more than a few minutes to answer all of them.
We chose to pay 0.03 USD for answering one page of questions.
This is a typical payment on \emph{CrowdFlower} for the contributors of the lowest level, and should maintain their commitment.
We chose to request answers from 20 distinct contributors to one question.
In the preliminary experiments, we requested only 3 answers, but in such a case, a single disagreement (e.g., when a contributor does not understand a question) makes the result unreliable. 
However, we did not want to increase the number too much, to keep the costs under control.

An instruction for a crowdsourcing experiment should be as simple as possible, yet answer all questions a contributor can ask.
Moreover, it must contain examples of real questions, both positive and negative, and all steps that should be undertaken to solve them.
For our experiment, we inform the contributors, that axioms are represented as sentences and their task is to decide whether a given sentence is true, false, or is not clear.
In the instructions, we also mentioned one true sentence, one false sentence, and one not clear sentence with an explanation in each case.
The full text of the instruction is also available in the \name{Git} repository.

\subsubsection{Experimental results}

For the crowdsourcing experiment, we generated two sets of axioms:  one with the minimal support threshold $\theta_\sigma=0.5$ and the other  with the threshold $\theta_\sigma=0.8$.
From each of the sets, we removed axioms that were logically entailed by the ontology or by the mined axioms for a superclass.
The first set was translated to 425 payable questions and 61 test questions.
Each of the payable questions was asked to 20 distinct contributors, leading to 8500 trusted answers.
The second set was translated to 168 payable questions and 56 test questions; the payable questions yielded $20\cdot 168=3360$ trusted answers.
Reliability of all contributors was checked with the test questions and when it dropped below 70\% (more than 30\% of the presented test questions had wrong answers), they were refused to continue answering the questions.

A summary of the results is presented in Table \ref{tab:cf-summary}.
In the first set, $69.41\%$ (resp. $38.69\%$ in the second set) of the axioms were accepted by at least $80\%$ of the contributors and $97.17\%$ (resp. $89.88\%$) by at least $55\%$ of the contributors.
Our aim was to verify whether SLDM can mine new, meaningful axioms that can be added to the ontology.
We found both results to answer positively to the question.
All the verified knowledge was new, because the axioms that could be inferred from the ontology were used as the test questions.


\begin{table}
\caption{A summary of the results of the crowdsourcing experiment.
For each question, we counted the numbers of \emph{Yes}, \emph{No}, and \emph{I don't know} answers and divided them into buckets: 0--5, 6--10, 11--15, and 16--20.
In the table, we report the number of questions having all three counts in the buckets given by the first three columns of the table.
In the last two columns specified are numbers of questions in the group, as an absolute number and relatively to the overall number of questions in the respective set.
For example, 3 questions (i.e., $0.71\%$ of all questions) in the first set were answered \emph{Yes} by 6 to 10 contributors, \emph{No} by 11 to 15 contributors and \emph{I don't know} by 0 to 5 contributors.
\label{tab:cf-summary}
}
\centering
\begin{tabular}{rr>{\centering\arraybackslash}p{1cm}|rr|rr}
\emph{Yes} & \emph{No} & \emph{I don't know} & \multicolumn{2}{c|}{$\theta_\sigma=0.5$} & \multicolumn{2}{c}{$\theta_\sigma=0.8$} \\
\hline
16--20 & 0--5 & 0--5 & 295 & $69.41\%$ & 65 & $38.69\%$ \\
11--15 & 6--10 & 0--5 & 40 & $9.41\%$ & 41 & $24.40\%$ \\
11--15 & 0--5 & 0--5 & 78 & $18.35\%$ & 45 & $26.79\%$ \\
6--10 & 11--15 & 0--5 & 3 & $0.71\%$ & 3 & $1.79\%$ \\
6--10 & 6--10 & 0--5 & 5 & $1.18\%$ & 13 & $7.74\%$ \\
0--5 & 16--20 & 0--5 & 2 & $0.47\%$ & 0 & $0.00\%$ \\
0--5 & 11--15 & 0--5 & 2 & $0.47\%$ & 1 & $0.60\%$ \\
\hline
\hline
\multicolumn{3}{c|}{overall} & 425 & $100.00\%$ & 168 & $100.00\%$ \\
\end{tabular}
\end{table}

To measure the level of disagreement for each question, we treated the numbers of \emph{Yes}, \emph{No}, and \emph{I don't know} answers as coordinates in a space and measured the Euclidean distance from all three crisp answers.
We treated the minimal distance as the disagreement measure, where higher value means higher disagreement.
For example, a question with 7 \emph{Yes}, 8 \emph{No}, and 5 \emph{I don't know} has coordinates $(7, 8, 5)$ and the nearest crisp answer is all \emph{No} with coordinates $(0, 20, 0)$ (distance: $\sqrt{218}$).
We present top 5 questions with the highest disagreement, selected from the axioms mined with the minimal support threshold $\theta_\sigma=0.5$ as given  in Table \ref{tab:cf-disagreement}.

Question 1 probably refers to a geographical position of a crater, but the name of the predicate is very vague and, as it is in the \texttt{dbp} namespace, it lacks a description.
Question 2 was mined, because $31,172$ out of $51,019$ instances, that is, over $50\%$, of \texttt{dbo:WrittenWork} is asserted to the class \texttt{http://purl.org/ontology/bibo/Book}.
Questions 3 and 4 display a similar problem with a complex structure and verbalization requiring knowledge about knowledge representation (KR).
Finally, question 5, on top of requiring knowledge about KR, requires also expert knowledge from the domain of law.

\begin{table*}
\caption{The top 5 questions with the most disagreement between the contributors.
Every question is presented with the axiom it originated from and the number of contributors which submitted a given answer.
For easier reading, the questions are numbered.
}
\label{tab:cf-disagreement}
\centering
\begin{tabular}{r|rr>{\centering\arraybackslash}p{1cm}|p{.75\textwidth}}
\# & \emph{Yes} & \emph{No} & \emph{I don't know} & A question and the axiom it originated from \\
\hline
1 & 7 & 8 & 5 & "Every crater has E or W. E or W is Literal." \\ 
& & & & \texttt{dbo:Crater \owlSubClassOf{} dbp:eOrW \owlSome{} rdf:PlainLiteral} \\
2 & 10 & 9 & 1 & "Every written work (Written work is any text written to read it (e.g. -  books, newspaper, articles)) is Book *." \\ 
& & & & \texttt{dbo:WrittenWork \owlSubClassOf{} <http://purl.org/ontology/bibo/Book>} \\
3 & 10 & 8 & 2 & "Every genre has instrument. instrument has is Primary Topic Of *. is Primary Topic Of * is Thing. instrument has label. label is Literal." \\ 
& & & & \texttt{dbo:Genre \owlSubClassOf{} dbo:instrument \owlSome{} ((foaf:isPrimaryTopicOf \owlSome{} owl:Thing) \owlAnd{} (rdfs:label \owlSome{} rdf:PlainLiteral))} \\
4 & 10 & 8 & 2 & "Every genre has instrument. instrument has is Primary Topic Of *. is Primary Topic Of * is Document *. is Primary Topic Of * has Language (A language of the resource.). Language (A language of the resource.) is string. is Primary Topic Of * has Language (A language of the resource.) which value is en." \\ 
& & & & \texttt{dbo:Genre \owlSubClassOf{} dbo:instrument \owlSome{} foaf:isPrimaryTopicOf \owlSome{} (foaf:Document \owlAnd{} dc:language \owlSome{} xsd:string \owlAnd{} dc:language \owlValue{} "en"\typedliteral{}xsd:string)} \\
5 & 10 & 6 & 4 & "Every journalist has nationality. nationality has Legislature. Legislature has see Also. see Also is Thing. Legislature has type that is Bicameralism *." \\ 
& & & & \texttt{dbo:Journalist \owlSubClassOf{} dbo:nationality \owlSome{} (dbp:legislature \owlSome{} ((rdfs:seeAlso \owlSome{} owl:Thing) \owlAnd{} (dbo:type \owlValue{} dbr:Bicameralism)))}
\end{tabular}
\end{table*}

\subsection{Extending the myExperiment ontology\label{sec:myexperiment}}

To provide to the reader an insight about how SLDM performs on a dataset different than \emph{DBpedia}, we performed an experimental evaluation on the RDF dataset published by myExperiment, a website designed for sharing and collaboration on scientific workflows describing experiments \cite{myexperiment}.
The dataset is published according to the Linked Data principles at \url{http://rdf.myexperiment.org/}.
It consists of 2,907,345 triples, describing 526,201 different subjects using 115 different predicates and 44 different types.
The used vocabulary is gathered in an ontology comprising  10 modules and described in a detailed way on the website\footnote{\url{http://rdf.myexperiment.org/ontologies/}}.
We used the RDF dataset as-is, without performing any logical reasoning on it.

We decided to run SLDM on every class that have at least 100 instances asserted to it in the dataset, that is, on 34 classes.
To keep the mined axioms simple and highly supported by the data, we set the minimal support threshold $\theta_\sigma$ to $0.9$, maximal depth to 1 and disabled sampling.
The full results are available in the \emph{Git} repository and  we provide a discussion of the mined patterns for the classes from the module Annotations below.

Table \ref{tab:my_experiment_annotations} presents these of the mined axioms, that were not logically entailed by the ontology.
All five of the considered classes exhibited the same pattern with respect to the usage of property \texttt{dct:hasFormat} (axioms 1, 3, 5, 6, 10).
While not presented in the description of the ontology, it seems that the property is used according to its intended usage\footnote{\url{http://dublincore.org/documents/dcmi-terms/#terms-hasFormat}}, that is, to link various representations of the same resource.
With respect to axiom 2, the documentation explains that this is a functional property to attach submission text to a submission.
The superclass in the axiom is Comment, and it seems reasonable that a comment has some text.
Axiom 4 correctly discovered part of the knowledge hidden only in a description of rating-score property stating that a rating score must be between 1 and 5 (the other part is already present in the ontology).
Axiom 7 is indirectly confirmed by the rest of the ontology, as the class Tagging seems to be reification of a ternary relation between an annotator, an annotated object and a tag.
Axiom 8 is represented in the ontology in a weaker form \texttt{dct:title \owlSome{} rdfs:Literal} and, arguably, the weaker form may be preferred by some ontology engineers, but in the dataset itself every time the predicate \texttt{dct:title} is used, the literal is of type \texttt{xsd:string}.
Finally, axiom 9 is not present in the documentation, but apparently it is used to link a tag to its HTML website.

This experiment shows that SLDM indeed can discover new facts about the data, those that are available only in textual form in the documentation, as well as those that reflect some, possibly unconscious, decisions about the design of a particular computer system handling the data.

\begin{table}
\caption{Axioms mined for the classes from the module Annotations of the myExperiment ontology.
Presented are only the axioms that are not logically entailed by the ontology.
The axioms are grouped by their subclass and numbered for easier reading.
}
\label{tab:my_experiment_annotations}
\begin{tabular}{rp{.8\columnwidth}}
\# & axiom \\
\hline
& \texttt{annotations:Comment} \owlSubClassOf{} \\
1 & \texttt{dct:hasFormat \owlSome{} owl:Thing} \\
2 & \texttt{base:text \owlSome{} xsd:string} \\
\hline
& \texttt{annotations:Favourite} \owlSubClassOf{} \\
3 & \texttt{dct:hasFormat \owlSome{} owl:Thing} \\
\hline
& \texttt{annotations:Rating} \owlSubClassOf{} \\
4 & \texttt{annotations:rating-score \owlSome{} xsd:decimal[<= 5]} \\
5 & \texttt{dct:hasFormat \owlSome{} owl:Thing} \\
\hline
& \texttt{annotations:Tagging} \owlSubClassOf{} \\
6 & \texttt{dct:hasFormat \owlSome{} owl:Thing} \\
7 & \texttt{annotations:uses-tag \owlSome{} owl:Thing} \\
\hline
& \texttt{annotations:Tag} \owlSubClassOf{} \\
8 & \texttt{dct:title \owlSome{} xsd:string} \\
9 & \texttt{foaf:homepage \owlSome{} owl:Thing} \\
10 & \texttt{dct:hasFormat \owlSome{} owl:Thing} \\
\hline
\end{tabular}
\end{table}


\subsection{Further experimental analysis of the algorithm\label{sec:further_exp}}

\subsubsection{Comparison for different random seeds}

In order to see how stable the results are depending on the choice of the random seed, we picked 10 different random seeds using \url{random.org} and performed the following experiment: we fixed the sample size to be 250, the minimal support threshold $\theta_\sigma$ to be $0.95$, and used the five classes from the bottom of the hierarchy depicted in Figure \ref{fig:hierarchy1}.
This way, for each of the classes, we obtained 10 sets of axioms.
For every axiom, we counted how many of the sets (extended with the \emph{DBpedia} ontology) entailed the axiom.
The detailed results are presented in Table \ref{tab:random_stability}.

In general, we obtained a very good stability, with at least 68\% of the axioms being mined for all of the random seeds.
In case of the ProgrammingLanguage class, every time exactly the same set of axioms was obtained.
For the Book class, 16 out of 18 axioms were entailed by all the sets and the remaining 2 axioms by 8 out of 10 sets.
Similar was the case of the Journalist class, but with, respectively, 11 and 5 out of 16 axioms.
Finally, for the Crater class, one axiom was entailed by 4 sets, and for the MusicGenre class, single axioms were entailed by 3, 4, and 6 sets.

\begin{table*}[t]
\caption{
Experimental analysis of the stability of the algorithm.
For each of the 5 classes, 10 sets of axioms were generated using different random seed.
The column \emph{distinct axioms} gives the number of distinct axioms mined, whereas the column \emph{all axioms} gives the number of all axioms in all of the mined sets.
The numbered columns correspond to the numbers of sets and the values in cell are the overlap size.
For example, value 5 in the column 8 of the row Journalist means that in case of 5 axioms, they were logically entailed by 8 out of 10 sets of the axioms.
These 8 sets may differ from axiom to axiom.
}
\label{tab:random_stability}
\centering
\begin{tabular}{l|rr|rrrrrrrrrrr}
class & distinct & all & \multicolumn{11}{c}{number of axioms entailed by given number of the sets} \\
& axioms & axioms & 1 & 2 & 3 & 4 & 5 & 6 & 7 & 8 & 9 & 10 & \\
\hline
Book & 18 & 160  &  0 & 0 & 0 & 0 & 0 & 0 & 0 & 2 & 0 & 16\\
Crater & 13 & 120  & 0 & 0 & 0 & 1 & 0 & 0 & 0 & 0 & 0 & 12\\
Journalist & 16 & 150  &  0 & 0 & 0 & 0 & 0 & 0 & 0 & 5 & 0 & 11\\
MusicGenre & 16 & 124  & 0 & 0 & 1 & 1 & 0 & 1 & 0 & 0 & 0 & 13\\
ProgrammingLanguage & 15 & 150   & 0 & 0 & 0 & 0 & 0 & 0 & 0 & 0 & 0 & 15\\
\end{tabular}
\end{table*}

\subsubsection{Comparison of different sampling strategies\label{sec:strategies_comparison}}
To gauge the variability of the obtained results with respect to the minimal support threshold $\theta_\sigma$, sample size, sampling strategy, and class, we executed the following experiment.
We selected all 5 classes from the bottom of the hierarchy presented in Figure \ref{fig:hierarchy1}; we considered the minimal support thresholds $\theta_\sigma$ from the set $\{0.5, 0.6, 0.7, 0.8, 0.9, 1.0\}$ and the sample sizes $n$ from the set $\{50, 100, 200, 500, 1000\}$.
We also considered all three sampling strategies.
This way we obtained $5\cdot 6\cdot 5\cdot 3=450$ configurations of SLDM.
For each of them, we executed SLDM, removed from the obtained set of axioms these that are logically entailed by the \emph{DBpedia} ontology, and counted the remaining ones.
The results are presented in Table \ref{tab:strategies}.

In 99/150, that is, $66\%$, of the cases, the predicates counting strategy yielded the highest number of mined axioms (ties included).
To further analyze the differences between the strategies, we wanted to see in how many cases all of the axioms mined using one strategy are logically entailed by the axioms mined with the other strategy.
The comparison presented in Table \ref{tab:strategies_comp} hints that the counting strategies generally perform better than the uniform strategy.
It is of no surprise as they both employ additional knowledge derived from computing statistics on the SPARQL endpoint.
The performance of both of the counting strategies is very similar, with a little bit of advantage for the predicates counting strategy.
Given that the SPARQL query (c.f. Section \ref{sec:index}) corresponding to the predicates counting strategy is simpler, probably this strategy should be favored over the triples counting strategy.

\begin{table}
\caption{Comparison of different sampling strategies: \emph{uni.} stands for uniform, \emph{pc} for predicates counting, \emph{tc} for triples counting.
A row corresponds to the set of axioms for which entailment is checked, while the columns to the set of axioms serving as the reference for the entailment checking.
For example, in 122 out of 150 cases (different classes, supports, and samples sizes) all of the axioms mined using the uniform strategy were all logically entailed by the axioms mined using the same settings, but with the predicates counting strategy.}
\label{tab:strategies_comp}
\centering
\begin{tabular}{l|lll}
 & \multicolumn{3}{c}{entailed by} \\
 & uni. & pc & tc \\
\hline
uni.  & -- & 122/150=$81\%$ & 121/150=$81\%$ \\
pc & 63/150=$42\%$ & -- & 100/150=$67\%$ \\
tc & 69/150=$46\%$ & 99/150=$66\%$ & -- \\ 
\end{tabular}
\end{table}

\begin{table}
\caption{Memory consumption of the whole Java Virtual Machine (JVM) while mining for the axioms for the \emph{DBpedia} class Book.
The reported values are maximal over 10 repeats for each 10 different random seeds and for each of 5 different sample sizes given a fixed minimal support threshold $\theta_\sigma$.
}
\label{tab:memory}
\centering
\begin{tabular}{l|rrrrrr}
$\theta_\sigma$ &  $0.5$ & $0.6$ & $0.7$ & $0.8$ & $0.9$ & $1.0$ \\
Memory [GB] & $9.82$ & $9.81$ & $5.44$ & $4.85$ & $2.64$ & $0.79$ \\

\end{tabular}
\end{table}

\begin{table*}
\caption{Number of axioms mined for a given class, using a given sampling strategy (\emph{uni.} stands for uniform, \emph{pc} for predicates counting, \emph{tc} for triples counting), sample size $n$ and minimal support threshold $\theta_\sigma$.
The axioms logically entailed by the \emph{DBpedia} ontology were removed from the count.
}
\label{tab:strategies}

\centering
\begin{tabular}{rr|rrr|rrr|rrr|rrr|rrrr}
 &  & \multicolumn{3}{c|}{Book} & \multicolumn{3}{c|}{Crater} & \multicolumn{3}{c|}{Journalist} & \multicolumn{3}{c|}{MusicGenre} & \multicolumn{4}{c}{ProgrammingLanguage} \\
$\theta_\sigma$ & $n$ & uni. & pc & tc & uni. & pc & tc & uni. & pc & tc & uni. & pc & tc & uni. & pc & tc \\
\hline
$0.5$ & 50 & 54 & 88 & 85 & 19 & 23 & 20 & 50 & 19 & 66 & 18 & 40 & 63 & 49 & 34 & 61 & \\
 & 100 & 87 & 102 & 87 & 22 & 23 & 21 & 69 & 19 & 64 & 13 & 35 & 50 & 32 & 57 & 51 & \\
 & 200 & 81 & 109 & 99 & 22 & 23 & 23 & 69 & 96 & 76 & 20 & 40 & 59 & 25 & 45 & 41 & \\
 & 500 & 96 & 106 & 102 & 20 & 20 & 20 & 96 & 77 & 96 & 20 & 34 & 47 & 22 & 37 & 38 & \\
 & 1000 & 96 & 111 & 113 & 20 & 20 & 20 & 113 & 125 & 113 & 22 & 30 & 31 & 22 & 22 & 22 & \\
\hline
$0.6$ & 50 & 58 & 61 & 60 & 16 & 21 & 20 & 38 & 79 & 57 & 7 & 25 & 30 & 23 & 31 & 28 & \\
 & 100 & 46 & 67 & 54 & 16 & 17 & 17 & 62 & 63 & 56 & 7 & 24 & 38 & 15 & 27 & 24 & \\
 & 200 & 51 & 66 & 61 & 16 & 16 & 17 & 61 & 65 & 60 & 9 & 29 & 34 & 13 & 20 & 23 & \\
 & 500 & 55 & 68 & 68 & 16 & 16 & 16 & 83 & 68 & 64 & 11 & 27 & 34 & 13 & 19 & 19 & \\
 & 1000 & 60 & 69 & 75 & 16 & 16 & 16 & 116 & 110 & 110 & 14 & 18 & 18 & 13 & 13 & 13 & \\
\hline
$0.7$ & 50 & 26 & 37 & 29 & 10 & 17 & 16 & 36 & 55 & 49 & 7 & 21 & 28 & 11 & 11 & 19 & \\
 & 100 & 39 & 42 & 41 & 7 & 16 & 10 & 53 & 55 & 58 & 7 & 17 & 29 & 11 & 16 & 15 & \\
 & 200 & 31 & 49 & 54 & 10 & 16 & 10 & 53 & 61 & 56 & 7 & 17 & 29 & 11 & 13 & 17 & \\
 & 500 & 41 & 52 & 55 & 10 & 10 & 10 & 85 & 83 & 71 & 7 & 13 & 23 & 11 & 11 & 13 & \\
 & 1000 & 40 & 53 & 54 & 10 & 10 & 10 & 105 & 105 & 91 & 8 & 9 & 10 & 11 & 11 & 11 & \\
\hline
$0.8$ & 50 & 14 & 29 & 17 & 7 & 10 & 7 & 34 & 61 & 16 & 7 & 11 & 17 & 9 & 11 & 13 & \\
 & 100 & 19 & 28 & 25 & 7 & 7 & 10 & 16 & 16 & 16 & 7 & 8 & 19 & 9 & 11 & 11 & \\
 & 200 & 21 & 26 & 25 & 7 & 7 & 7 & 16 & 16 & 16 & 7 & 8 & 18 & 9 & 11 & 11 & \\
 & 500 & 26 & 27 & 31 & 7 & 7 & 7 & 41 & 26 & 17 & 7 & 7 & 16 & 11 & 11 & 11 & \\
 & 1000 & 24 & 36 & 27 & 7 & 7 & 7 & 85 & 78 & 32 & 8 & 8 & 8 & 10 & 10 & 10 & \\
\hline
$0.9$ & 50 & 10 & 14 & 9 & 7 & 7 & 7 & 15 & 15 & 15 & 7 & 7 & 15 & 9 & 11 & 11 & \\
 & 100 & 9 & 17 & 15 & 7 & 7 & 7 & 13 & 15 & 15 & 7 & 8 & 11 & 9 & 11 & 9 & \\
 & 200 & 11 & 12 & 10 & 7 & 7 & 7 & 13 & 15 & 15 & 7 & 7 & 13 & 9 & 11 & 11 & \\
 & 500 & 9 & 12 & 14 & 7 & 7 & 7 & 15 & 15 & 15 & 8 & 8 & 8 & 9 & 10 & 9 & \\
 & 1000 & 10 & 12 & 12 & 7 & 7 & 7 & 15 & 15 & 15 & 8 & 8 & 8 & 9 & 9 & 9 & \\
\hline
$1.0$ & 50 & 7 & 9 & 9 & 7 & 7 & 7 & 3 & 14 & 13 & 6 & 6 & 6 & 8 & 5 & 6 & \\
 & 100 & 4 & 5 & 6 & 7 & 5 & 6 & 3 & 11 & 3 & 6 & 5 & 5 & 7 & 7 & 5 & \\
 & 200 & 2 & 6 & 4 & 5 & 5 & 5 & 3 & 3 & 3 & 1 & 3 & 5 & 5 & 7 & 5 & \\
 & 500 & 2 & 2 & 2 & 1 & 1 & 1 & 3 & 3 & 3 & 1 & 5 & 5 & 5 & 5 & 5 & \\
 & 1000 & 2 & 2 & 2 & 1 & 1 & 1 & 3 & 3 & 3 & 1 & 3 & 3 & 0 & 0 & 0 & \\
\hline

\end{tabular}

\end{table*}

\subsubsection{Runtime performance}

To judge the performance of the implementation described in Section \ref{sec:protege}, we mined axioms for the \emph{DBpedia} class Book using various settings.
We used uniform sampling strategy and randomly selected 10 different random seeds to account for different samples.
We considered each of the following 6 minimal support thresholds $\theta_\sigma$: $\{0.5, 0.6, 0.7, 0.8, 0.9, 1.0\}$ and each of the following 5 sample sizes: $\{50, 100, 200, 500, 1000\}$, obtaining $10\cdot 6\cdot 5=300$ different settings.
Moreover, each of the experiments was repeated 10 times to account for the variability normal to the computer system, yielding overall $3,000$ experiments being run in the timespan of 15 hours.
The experiments were run on a server equipped with two \emph{Intel Xeon E5-2630 v3} running at $2.4$ GHz each and with 256 GB of the RAM.
The SPARQL endpoint was set up on the same machine.

We measured overall CPU time of the mining, including starting of the Java Virtual Machine (JVM) and including the system CPU time accounted to the process.
A single run of the experiment never took more than 2 minutes, even for the lowest minimal support threshold and the highest sample size.
The detailed statistics are presented in Figure \ref{fig:cputime}.

While performing the experiment, we also measured maximal memory consumption of the JVM.
As garbage collection is one of the features in \emph{Java}, we report only the highest values in Table \ref{tab:memory}.
Without detailed profiling of the execution, it is unclear if, and to what extent, these could be lowered.

Finally, during the experiment, we counted the number of queries posed to the SPARQL endpoint.
The number never exceeded 400 and the detailed statistics, aggregated in the same manner as for CPU time, are presented in Figure \ref{fig:queries_posed}.

\begin{figure*}
\centering
\resizebox{!}{.43\textheight}
{
\input{cputime.pgf}
}
\caption{CPU time used by SLDM for mining axioms for the Book class in \emph{DBpedia}, averaged across 10 repeats for each of 10 different random seeds.
Each chart corresponds to a fixed minimal support threshold $\theta_\sigma$.
The charts all share the same horizontal axis presenting the sample size.
The vertical axes are shared within rows and present the overall CPU time of the mining (including starting of Java Virtual Machine (JVM)).
}
\label{fig:cputime}

\centering
\resizebox{!}{.43\textheight}
{
\input{sparql.pgf}
}
\caption{
Number of SPARQL queries posed to the SPARQL endpoint by SLDM for mining axioms for the Book class in \emph{DBpedia}, averaged across 10 repeats for each of 10 different random seeds.
Each chart corresponds to a fixed minimal support threshold $\theta_\sigma$.
The charts all share the same horizontal axis presenting the sample size and the vertical axes 
are shared within rows.
}
\label{fig:queries_posed}
\end{figure*}



\section{Conclusions}
In this paper, we presented SLDM, a novel approach for mining ontological axioms directly from online RDF datasets.
Capabilities of SLDM cover the whole grammar of OWL 2 EL superclass expressions.
SLDM is readily available for use as a \protege{} plugin available for download from the \name{Git} repository \url{https://bitbucket.org/jpotoniec/sldm}.

We also showed that SLDM can be used to mine RDF Data Shapes instead of ontological axioms.
We presented a transformation from SLDM axioms to the corresponding shapes expressed in Shapes Constraint Language (\shacl).

To validate that SLDM can be applied in a real use-case, we mined axioms for a set of classes from the \name{DBpedia} ontology, and conducted a crowdsourcing experiment to validate them.
The experiment confirmed our hypothesis, as most of the axioms was accepted by the contributors of a crowdsourcing platform.
All materials used for the experiment and the obtained results are published to the aforementioned  \name{Git} repository.
We also show that the required computational resources are in very reasonable limits and are easy to provide using contemporary computers.

In the future, we would like to cover more expressive variants of OWL 2, for example, by including universal quantification (\owl{only}).
We also would like to extend SLDM to mine more types of axioms, for example, to express disjointness.

\paragraph{Acknowledgements} 
Jedrzej Potoniec acknowledges the support received from the Polish National Science Center (Grant No 2013/11/N/ST6/03065).
This work was partially supported by the PARENT-BRIDGE program of Foundation for Polish Science, co-financed by the  European Union, Regional Development Fund (Grant No POMOST/2013-7/8).
Agnieszka \L{}awrynowicz acknowledges the support from the Polish National Science Center (Grant No 2014/13/D/ST6/02076).







\bibliographystyle{elsarticle-harv} 
\bibliography{jws,references}


%
%
%
\end{document}